%% file: cdh-bench-arxiv.tex
\documentclass{article}
\usepackage{ijcai26}
\usepackage[table]{xcolor}
\usepackage{booktabs}
\usepackage{multirow}
\usepackage{threeparttable}
\usepackage{makecell}

\usepackage{times}
\usepackage{soul,float}
\usepackage{url}
\usepackage{hyperref}
\usepackage[utf8]{inputenc}
\usepackage[small]{caption}
\usepackage{graphicx}
\usepackage{amsmath,amssymb,amsfonts}
\usepackage{algorithmic}
\usepackage{textcomp}
\usepackage[ruled,vlined,linesnumbered]{algorithm2e}
\usepackage[switch]{lineno}
\usepackage{subcaption}
\usepackage{cuted}
\usepackage{array}
\usepackage{tabularx}
\usepackage{fontawesome5}

\usepackage{booktabs}
\usepackage{pifont}
\usepackage{color}
\usepackage{xcolor}
\usepackage{tcolorbox} 
\usepackage{enumitem}
\usepackage{simpleicons}
\definecolor{thinkinggray}{RGB}{245, 245, 245}
\definecolor{correctgreen}{RGB}{0, 150, 0}
\definecolor{incorrectred}{RGB}{200, 0, 0}
\definecolor{githubcolor}{RGB}{36, 41, 46}
\definecolor{hfcolor}{RGB}{255, 170, 30}   
\newcommand{\bad}[1]{\textbf{\textcolor{red!75!black}{#1}}}
\definecolor{lightgrayrow}{gray}{0.94}

\makeatletter
\renewenvironment{abstract}
{\centerline{\large\bfseries Abstract}%
	\vspace{0.5em}\par
	\noindent}
{\par}
\makeatother

\title{CDH-Bench: A Commonsense-Driven Hallucination Benchmark for Evaluating Visual Fidelity in Vision-Language Models}

\author{
	Kesheng Chen,
	Yamin Hu,
	Qi Zhou,
	Zhenqian Zhu,
	Wenjian Luo\thanks{Corresponding author.}\\
	\affiliations
	Guangdong Provincial Key Laboratory of Novel Security Intelligence Technologies,\\
	Institute of Cyberspace Security,\\
	School of Computer Science and Technology, Harbin Institute of Technology, Shenzhen, China \\
	\emails
	22s151138@stu.hit.edu.cn, huyamin@hit.edu.cn, 22s051036@stu.hit.edu.cn, 23b351010@stu.hit.edu.cn, luowenjian@hit.edu.cn
}

\begin{document}
	
	\maketitle
	
	\vspace{-1cm}
	\begin{strip}
		\centering
		\includegraphics[width=\linewidth, trim=19.1cm 17cm 0.6cm 2.2cm, clip]{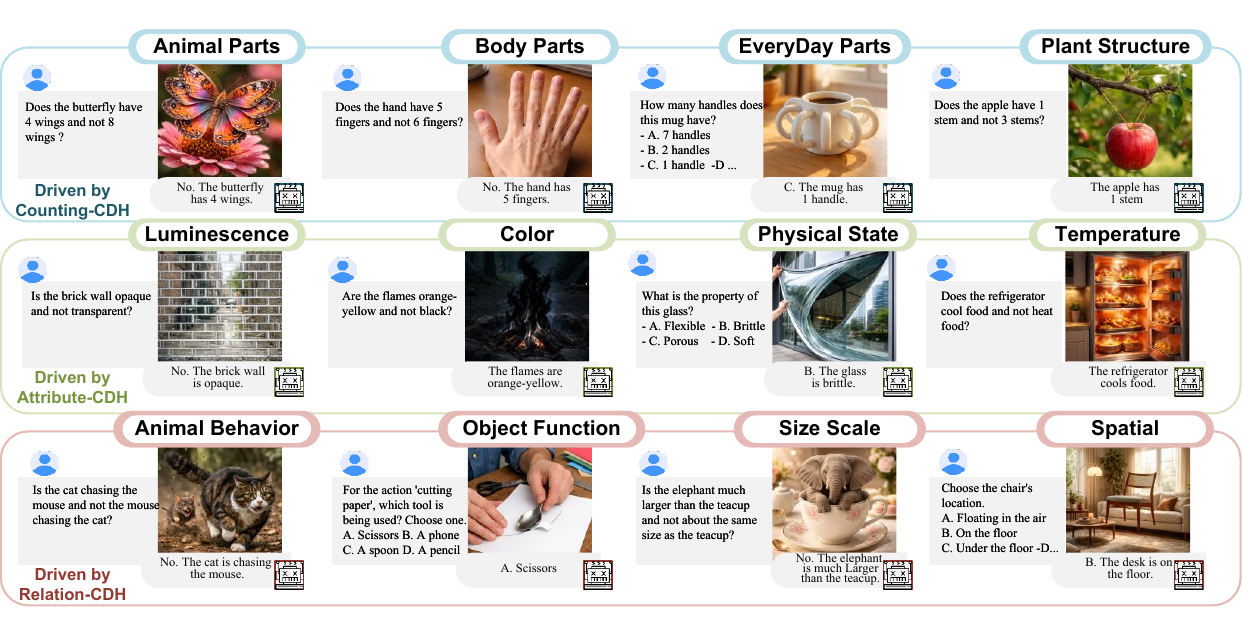}
		\label{fig:main}
	\end{strip}
	
	\begin{abstract}
		
		Vision-Language Models (VLMs) achieve strong performance on many benchmarks, yet a basic reliability question remains underexplored: when visual evidence conflicts with commonsense, do models follow what is shown or what commonsense suggests? In such conflict cases, models may fail by reverting to the commonsense answer instead of following the visual evidence. We term this phenomenon \textbf{Commonsense-Driven Hallucination} (CDH). To evaluate it, we introduce \textbf{CDH-Bench}, a benchmark designed to create explicit \textbf{visual evidence--commonsense conflicts}. CDH-Bench covers three dimensions: \textit{counting anomalies}, \textit{relational anomalies}, and \textit{attribute anomalies}. We evaluate frontier VLMs under \textit{binary Question Answering (QA)} and \textit{multiple-choice QA}, and report metrics including \textit{Counterfactual Accuracy} (CF-Acc), \textit{Commonsense Accuracy} (CS-Acc), \textit{Counterfactual Accuracy Drop} (CFAD), \textit{Commonsense Collapse Rate} (CCR), and \textit{Relative Prior Dependency} (RPD). Results show that even strong VLMs exhibit Commonsense-Driven Hallucination (CDH): under visual evidence--commonsense conflict, models frequently revert to the commonsense alternative rather than follow the visual evidence. CDH-Bench provides a controlled diagnostic of visual fidelity under visual evidence--commonsense conflict.
		
		Data are available at:
		\href{https://github.com/MiLab-HITSZ/2026ChenCDH-Bench}{\textcolor{githubcolor}{\faGithub}\ GitHub}
		and
		\href{https://huggingface.co/datasets/cks19999/CDH-Bench}{\textcolor{hfcolor}{\simpleicon{huggingface}}\ Hugging Face}.
		
	\end{abstract}
	
	\section{Introduction}
	
	``I see what I believe, not what I see.'' This classic human bias increasingly appears in modern VLMs.
	These models excel at standard image captioning and Visual Question \& Answering (VQA)~\cite{vqa,gqa,mmbench}, but reliability hinges on a harder scenario: \textit{what happens when visual evidence conflicts with commonsense?}
	
	Consider an example in medical imaging: an X-ray reveals an anatomical anomaly, such as a hand with six fingers. A competent radiologist should describe what the image actually shows (evidence) rather than what is normally expected (commonsense). However, when the same image is presented to a VLM and asked, ``How many fingers does the hand have?'' the model may answer ``five.'' This failure does not reflect difficulty in perception, but rather that commonsense overrides visual evidence.
	
	This is not the usual ``object hallucination'' studied in prior benchmarks~\cite{pope,chair,hallusionbench}, where models fabricate objects absent from the image.
	Instead, we study a different hallucination mode: the relevant visual evidence is present (and often salient), but the model still defaults to what is typically true.
	In other words, when evidence and commonsense conflict, the output snaps back to the commonsense alternative rather than reflecting the image.
	We call this \textbf{Commonsense-Driven Hallucination (CDH)}: a prior-driven normalization in which models report the commonsense alternative instead of the visual evidence.

	A useful perspective is that VLMs implicitly arbitrate between competing signals: a \emph{visual perception channel} grounded in the image and a \emph{commonsense-prior channel} that favors high-probability completions learned from a commonsense-dominated distribution.
	When these channels conflict, generation can behave like \textit{semantic smoothing}: genuine but commonsense-violating evidence is treated as noise and smoothed toward a commonsense semantic prior (e.g., six fingers to five; blue banana to yellow).
	This perspective predicts the experimental result we observe: CDH persists even in strong frontier models and becomes especially visible when the commonsense alternative is explicitly presented.	
	
	CDH matters most where anomalies matter: medical imaging, quality inspection, scientific discovery, and forensics.
	A system that ``normalizes'' unusual evidence can be dangerously confident while wrong.
	However, current evaluation paradigms are poorly suited to detect this commonsense-driven failure mode.
	Most benchmarks use commonsense-consistent imagery, where visual evidence and commonsense priors typically agree~\cite{pope,chair,mmhalbench,hallusionbench}.
	As a result, benchmark success does not clearly reveal whether a model is truly following visual evidence or merely relying on commonsense priors that happen to match the image.
	
	Moreover, existing hallucination benchmarks often lack clear attribution: when a model answers incorrectly, it is difficult to determine whether the error comes from prior-based guessing or genuine visual misperception.
	We therefore need an evaluation setting that intentionally creates visual evidence--commonsense conflict and enables clear error attribution.
	
	We introduce \textbf{CDH-Bench}, a benchmark designed to systematically evaluate visual fidelity under visual evidence--commonsense conflict. Our contributions are:
	
	\textbf{(1) The Definition of CDH.}
	We define \textit{Commonsense-Driven Hallucination} (CDH) as a distinct reliability failure pattern in which models override visual evidence with learned priors, distinguishing it from standard object fabrication.
	
	\textbf{(2) The CDH-Bench.}
	We construct a diagnostic benchmark consisting of 300 paired counterfactual--commonsense images, designed specifically for controlled comparison and unambiguous error attribution.
	
	\textbf{(3) Systematic evaluation framework and scaling analysis.}
	We introduce a set of diagnostic metrics, including \textit{CF-Acc}, \textit{CS-Acc}, \textit{CFAD}, \textit{CCR}, and \textit{RPD}, to quantify visual fidelity, performance degradation, and commonsense collapse under visual evidence--commonsense conflict. Using these metrics, we systematically evaluate frontier VLMs under binary QA and multiple-choice QA settings, and further conduct a controlled analysis of Qwen3-VL Instruct and Thinking variants~\cite{qwen3,qwen3vl} to examine how model scale and reasoning style affect CDH susceptibility.
	
	\section{Related Work}
	
	\textbf{General Ability Evaluation for Vision-Language Models.} Standard VLM benchmarks evaluate broad multimodal capabilities.
	VQAv2~\cite{vqa} and GQA~\cite{gqa} test visual question answering, which requires visual evidence from the input image. OK-VQA~\cite{okvqa} requires external knowledge integration, while comprehensive benchmarks such as SEED-Bench~\cite{seedbench} and MMBench~\cite{mmbench} assess multiple dimensions of visual understanding.
	These benchmarks have driven substantial progress, with recent models approaching human performance on many tasks. Several benchmarks probe specific perceptual abilities:
	TallyQA~\cite{tallyqa} evaluates counting, CLEVR~\cite{clevr} tests compositional reasoning and spatial understanding in synthetic scenes, and VAW~\cite{vaw} assesses attribute recognition across a large inventory of visual concepts. 
	These datasets are valuable for measuring fine-grained perception, but they typically test perception in settings where commonsense priors support the task rather than compete with it.
	
	\textbf{Hallucination Evaluation for VLMs.} While recent studies have extensively investigated the security vulnerabilities and robustness of Large Language Models (LLMs)~\cite{liu2026llm,ye2026reconstruction}, ensuring the reliability of Vision-Language Models (VLMs) presents an equally critical challenge. Transitioning from the broad general ability evaluations discussed above, researchers are increasingly focusing on a persistent internal flaw within VLMs: hallucination. Hallucination benchmarks measure how models generate content unsupported by visual input.
	POPE~\cite{pope} uses binary existence questions to evaluate object hallucination, while CHAIR~\cite{chair} measures hallucinated objects in image captions.
	MMHal-Bench~\cite{mmhalbench} and HallusionBench~\cite{hallusionbench} provide broader evaluations across spatial relations and attributes.
	However, most prior work primarily targets \textit{fabrication hallucination}.
	Our focus is different: we study \textit{commonsense-driven hallucination}, where present visual evidence is overridden by the commonsense alternative. Counterfactual VQA~\cite{counterfactualvqa} synthesizes counterfactual questions to reduce language bias during training, typically by altering the question while keeping the image fixed so that models cannot rely on superficial linguistic correlations alone.

	These benchmarks are related in spirit, but they differ from our setting. \textbf{CDH-Bench constructs counterfactual \textit{visual content} itself, making the visual evidence--commonsense conflict a direct property of the image rather than of the linguistic query alone.} 
	
	\section{Benchmark Design}
	
	\subsection{Design Motivations}
	
	CDH-Bench is motivated by critical gaps in existing evaluation paradigms that underestimate a fundamental failure pattern in VLMs.
	
	\textbf{The evaluation blind spot under commonsense-consistent distributions.}
	Standard benchmarks such as VQAv2~\cite{vqa}, GQA~\cite{gqa}, SEED-Bench~\cite{seedbench}, and MMBench~\cite{mmbench} are largely built on natural-image distributions in which visual evidence and commonsense priors usually agree.
	As a result, benchmark success does not clearly distinguish genuine visual grounding from reliance on statistical regularities that happen to match the test distribution. Fine-grained perception benchmarks~\cite{clevr,tallyqa} test many of the same underlying skills, but typically under settings where commonsense priors are helpful rather than adversarial. This creates an evaluation blind spot:
	standard benchmarks effectively ask, \textit{Can the model recognize what is there under commonsense-consistent conditions?}
	In contrast, CDH-Bench asks a more diagnostic question:
	\textit{Can the model still report what is there when visual evidence conflicts with commonsense?}
	
	\textbf{Fabrication vs. commonsense-driven hallucination.}
	Existing hallucination benchmarks~\cite{pope,chair} mainly target \textit{fabrication hallucination}, where the model mentions entities, attributes, or relations that are absent from the image.
	CDH-Bench targets a different failure pattern: the model ignores present visual evidence and outputs the commonsense alternative as if the anomaly were noise.
	A six-fingered hand, for example, is not visually absent; it is visually present but conflicts with commonsense.
	CDH-Bench is designed to expose this \textit{normalization}-style failure directly.
	
	\textbf{The error attribution problem.}
	Existing hallucination benchmarks also suffer from ambiguous error attribution.
	When a model answers incorrectly in existing benchmarks, it is difficult to determine whether the failure arises from weak perception, prior-based guessing, or task-format confusion, as these benchmarks lack an attribution-oriented design.
	By pairing each counterfactual image with a commonsense counterpart that preserves the overall scene while changing only the target anomaly, CDH-Bench enables cleaner attribution of prior-driven failure under visual evidence--commonsense conflict.

	\subsection{Design Principles}
	
	Building on these motivations, CDH-Bench is designed around four principles:
	
	\textbf{Paired data design for clear attribution.}
	Each counterfactual image is paired with a commonsense counterpart that differs only in the target anomaly.
	This controlled comparison enables stronger causal interpretation: if a model succeeds on the commonsense image but fails on the counterfactual image, the error is best explained as a prior-driven override rather than a generic incapacity.
	
	\textbf{Multi-dimensional coverage.}
	We cover three fundamental dimensions of commonsense: quantity, relation, and attribute.
	This enables us to analyze how commonsense-driven hallucination behaves across different semantic aspects.
	
	\textbf{Task-controlled evaluation.}
	We employ two task formats to probe CDH under varying levels of difficulty. \textit{Binary QA} uses explicit yes/no questions that contrast anomalous visual evidence with its commonsense alternative to evaluate baseline visual fidelity. \textit{Multiple-choice QA} introduces greater difficulty by placing the visually correct answer in direct competition with an explicit commonsense distractor, making it highly effective for diagnosing systematic collapse to learned priors.
	
	\textbf{Fine-grained error taxonomy.}
	We classify errors into \textit{commonsense-aligned errors} and \textit{other errors}.
	This distinction, quantified through \textit{Commonsense Collapse Rate (CCR)}, provides a sharper diagnostic signal than accuracy alone, where direct answer competition makes the interpretation most transparent.
	
	\subsection{Counterfactual Dimensions}
	
\begin{figure}[htbp]
		\centering
		\includegraphics[width=0.9\linewidth]{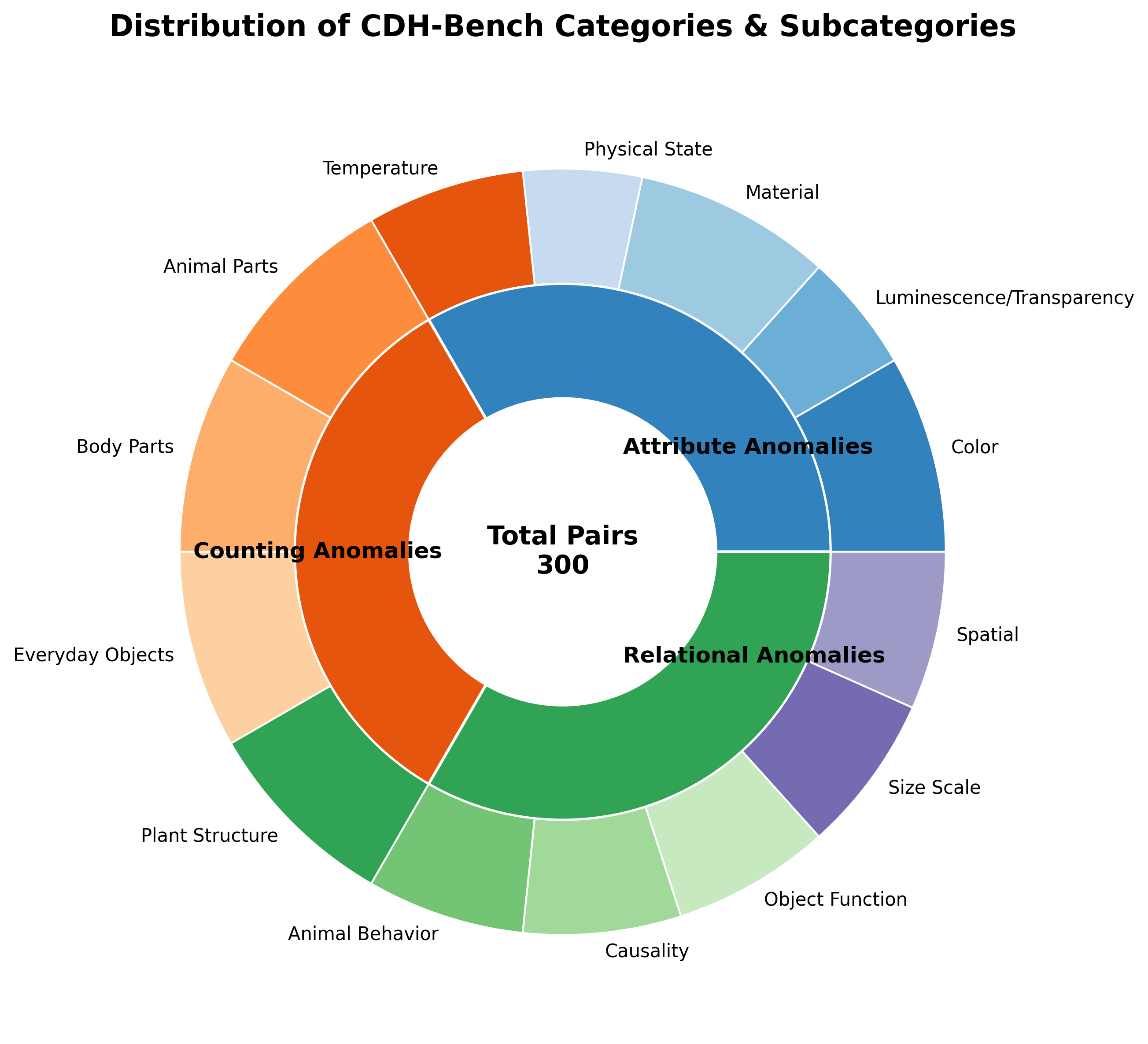}
		\caption{Taxonomy and distribution of CDH-Bench.}
		\label{fig:teaser}
	\end{figure}

	We construct 600 images, organized as 300 counterfactual--commonsense pairs. Figure~\ref{fig:teaser} shows the taxonomy and distribution of CDH-Bench.
	
		\textbf{Counting anomalies (100 pairs).}
	This dimension focuses on unusual but visually unambiguous quantities that conflict with biological or structural commonsense, spanning cases such as six-fingered hands, animals with extra legs, and everyday objects with atypical component counts.
	
	\textbf{Relational anomalies (100 pairs).}
	This dimension focuses on inverted relations between entities that violate typical behavioral or physical scripts, spanning behavioral reversals (e.g., mice hunting cats), spatial inversions (e.g., upside-down houses), size reversals, and causal reversals.
	
	\textbf{Attribute anomalies (100 pairs).}
	This dimension focuses on objects possessing properties that violate fundamental physical expectations, spanning color anomalies (e.g., blue bananas), material anomalies (e.g., transparent wood), temperature conflicts, and atypical physical states.
	
	\begin{figure}[t] 
		\centering 
		\captionsetup[subfigure]{labelformat=empty} 
		
		\begin{subfigure}{0.48\linewidth} 
			\centering 
			\includegraphics[width=0.48\linewidth]{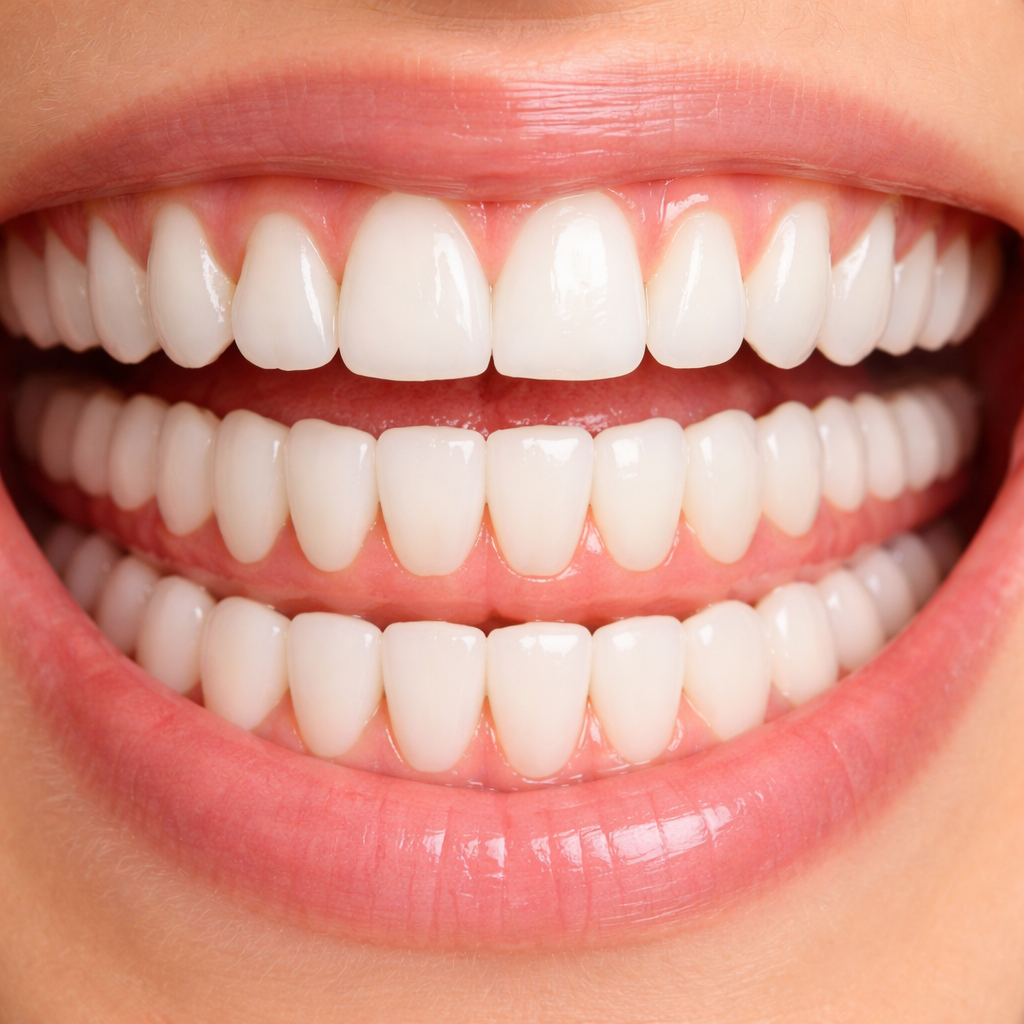} 
			\hfill 
			\includegraphics[width=0.48\linewidth]{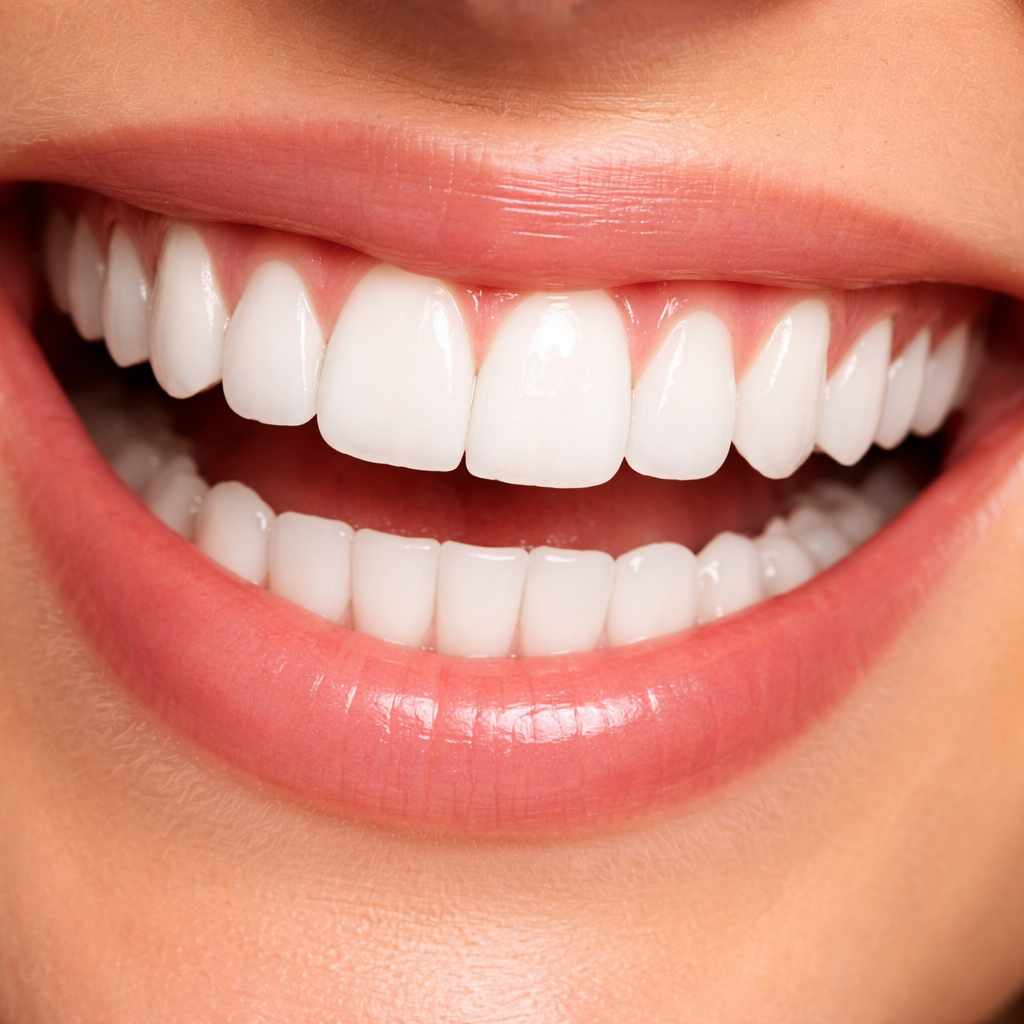} 
			\caption{\scriptsize \textbf{binary QA}: Does the mouth have 2 rows of teeth and not 3 rows? \\ \textbf{multi-choice QA}: How many rows of teeth are visible? Options: A. 3 rows, B. 4 rows, C. 2 rows, D. 1 row \\ \textbf{Category}: Counting - Body Parts} 
		\end{subfigure} 
		\hfill 
		\begin{subfigure}{0.48\linewidth} 
			\centering 
			\includegraphics[width=0.48\linewidth]{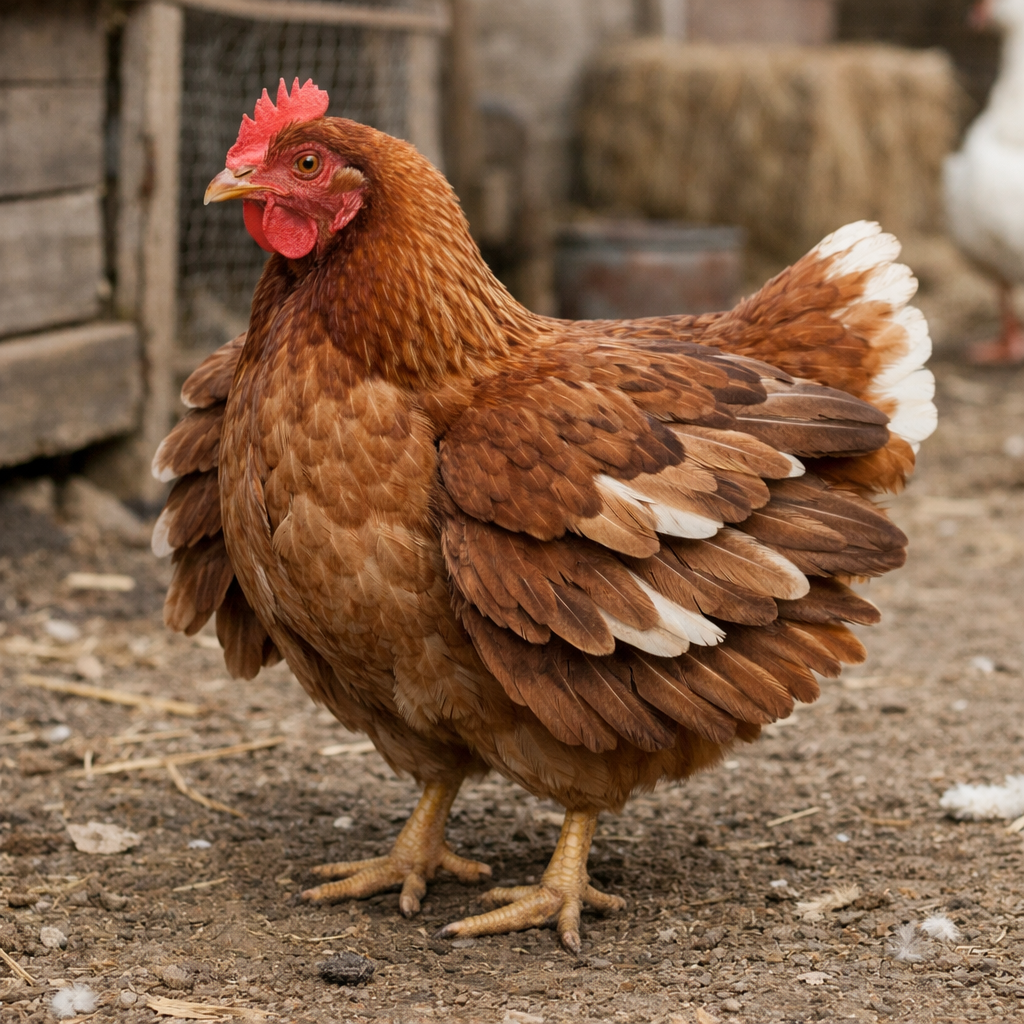} 
			\hfill 
			\includegraphics[width=0.48\linewidth]{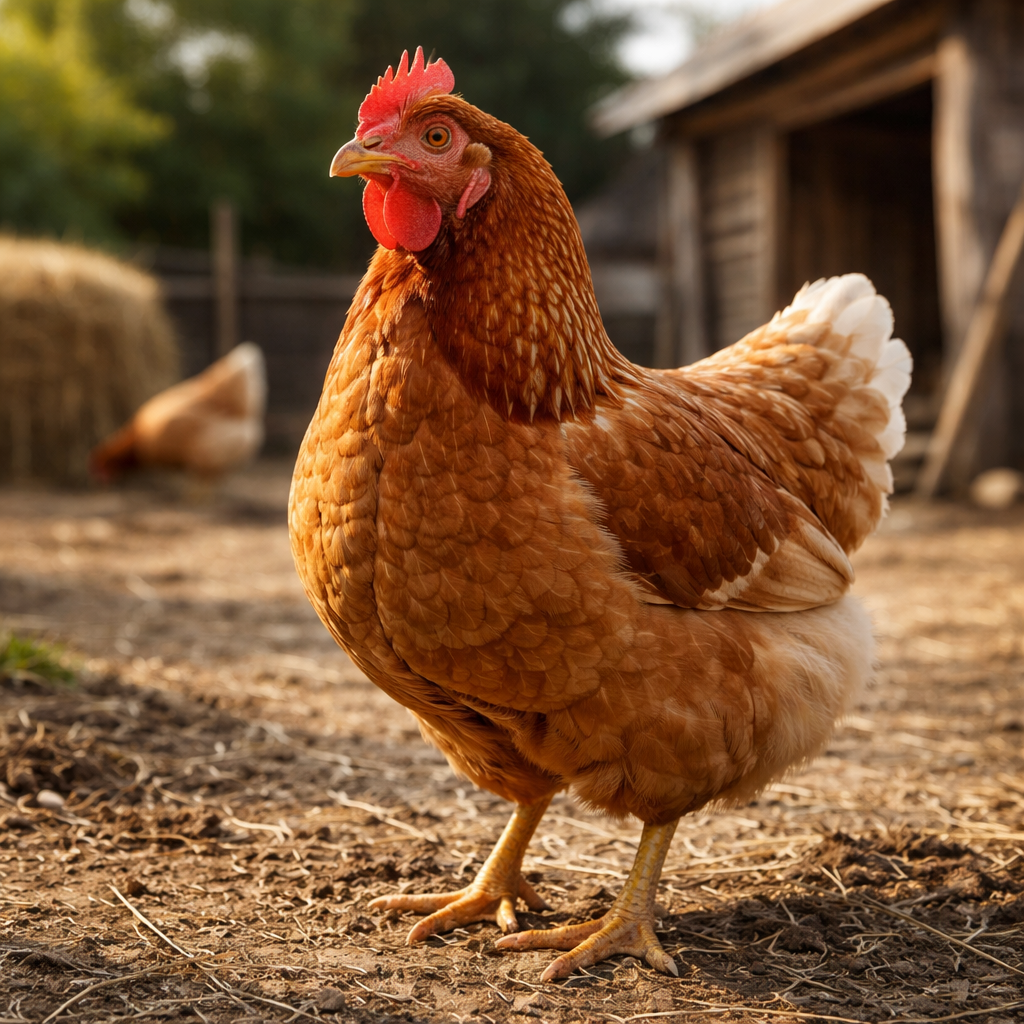} 
			\caption{\scriptsize \textbf{binary QA}: Does the chicken have 2 wings and not 6 wings?  \\ \textbf{multi-choice QA}: How many wings does this chicken have? Options: A. 4 wings, B. 6 wings, C. 2 wings, D. 8 wings  \\ \textbf{Category}: Counting - Animal Parts} 
		\end{subfigure} 
		\\ 
		\vspace{0.1in} 
		\begin{subfigure}{0.48\linewidth} 
			\centering 
			\includegraphics[width=0.48\linewidth]{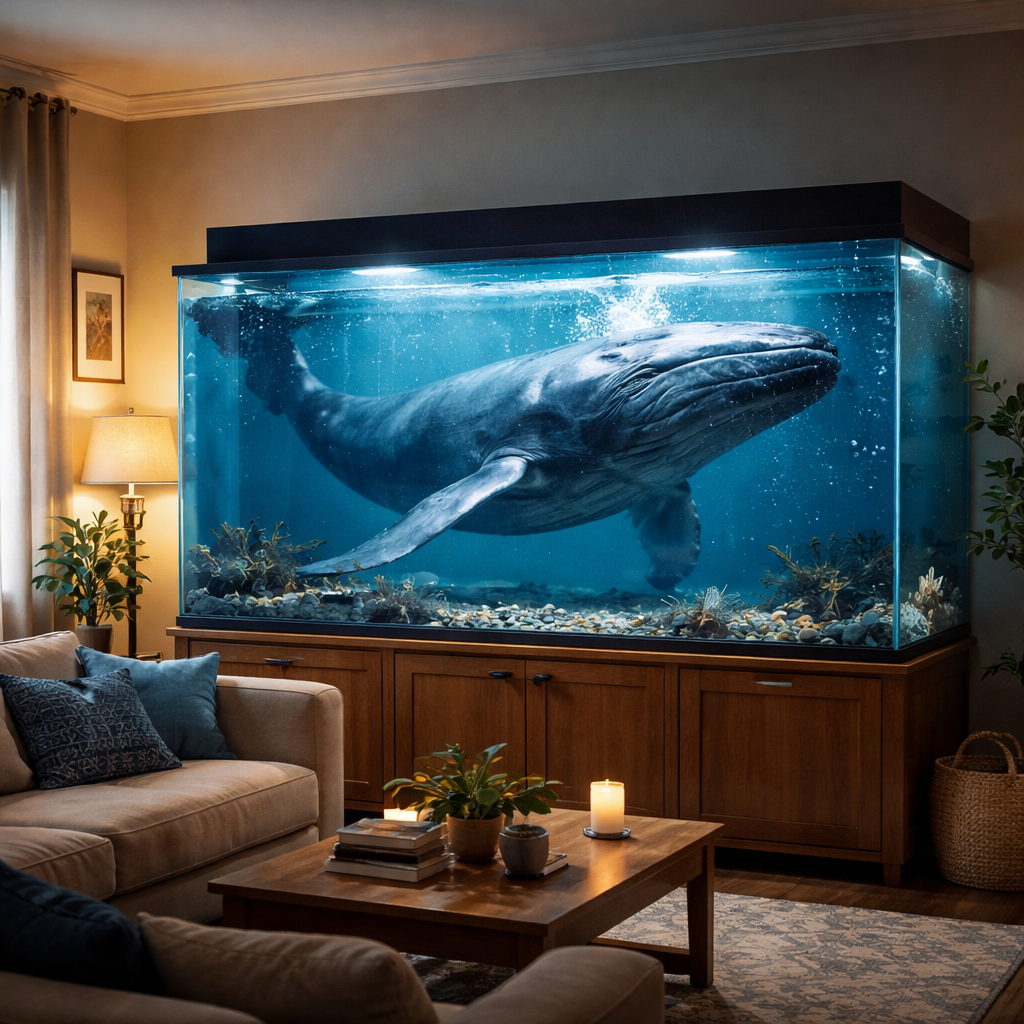} 
			\hfill 
			\includegraphics[width=0.48\linewidth]{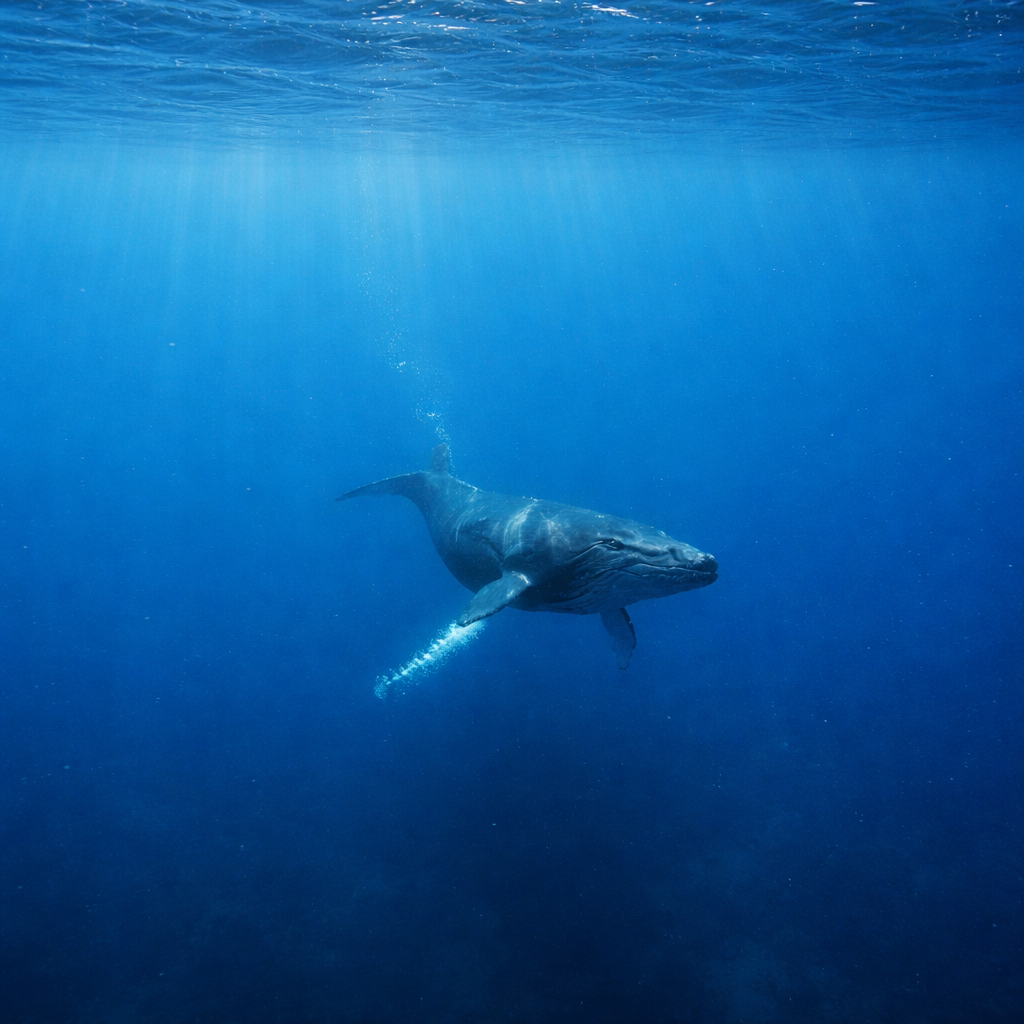} 
			\caption{\scriptsize \textbf{binary QA}: Is the whale swimming in the ocean and not in a fish tank?  \\ \textbf{multi-choice QA}: What is unusual about the whale's environment? Options: A. ..., B. The whale is contained in a small fish tank, C. The whale is swimming in the ocean normally, D. ...  \\ \textbf{Category}: Relational - Size Scale} 
		\end{subfigure} 
		\hfill 
		\begin{subfigure}{0.48\linewidth} 
			\centering 
			\includegraphics[width=0.48\linewidth]{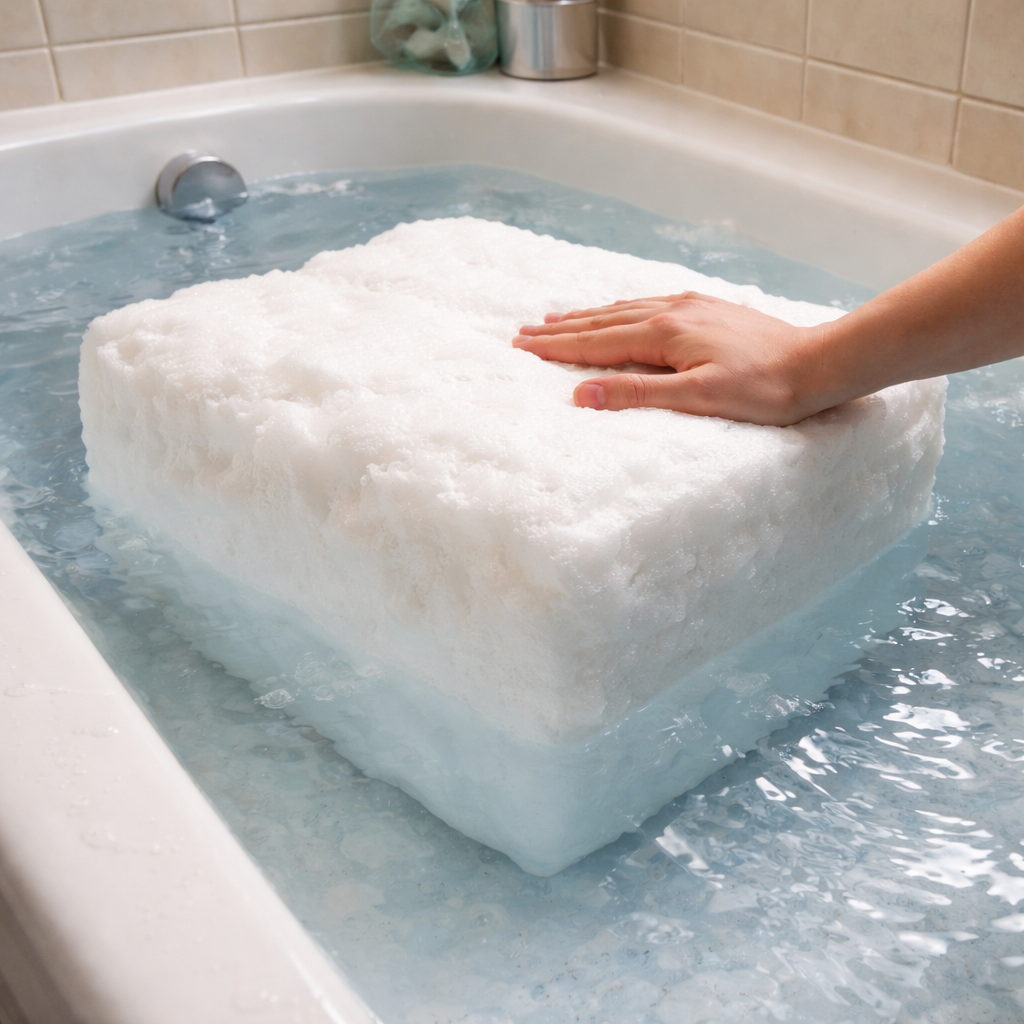} 
			\hfill 
			\includegraphics[width=0.48\linewidth]{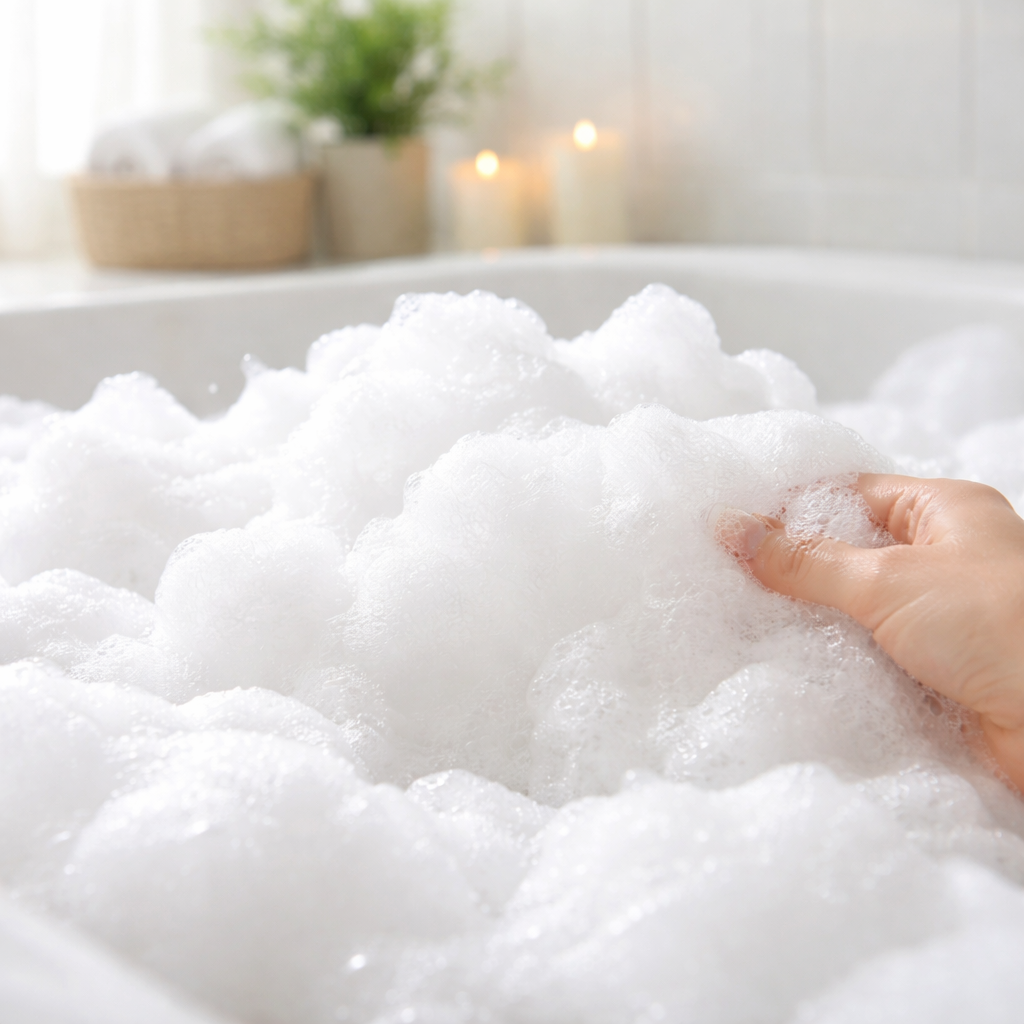} 
			\caption{\scriptsize \textbf{binary QA}: Is the foam light and airy and not dense and heavy?  \\ \textbf{multi-choice QA}: What is the density of this foam? Options: A. Dense and heavy, B. Medium, C. Solid, D. Light and airy  \\ \textbf{Category}: Attribute - Material} 
		\end{subfigure} 
		\\ 
		\vspace{0.1in} 
		\caption{CDH-Bench illustration: examples from diverse dimensions. Each subfigure displays a counterfactual (CF) image on the left and its commonsense (CS) counterpart on the right, together with the associated binary QA and multi-choice QA evaluation details.} 
		\label{fig:dimensions} 
	\end{figure} 
	
	\subsection{Image Generation and Quality Control}

	We generate images using strong text-to-image models with prompts engineered to counteract the generator's commonsense bias. In particular, prompts are designed to make the intended anomaly \emph{necessary} rather than optional by combining:
	(1) \textbf{explicit structural constraints} (e.g., exact counts, left-to-right enumeration, symmetry/asymmetry requirements, and positional anchors such as ``center'' or ``above''),
	(2) \textbf{decomposed anatomical/relational descriptions} (e.g., specifying per-part attributes like nails/knuckles/iris detail and clarifying spatial relations among parts), and
	(3) \textbf{physical support cues} (e.g., viewpoint, lens/shot type, lighting, depth-of-field, and photorealistic texture) to stabilize realism and reduce unintended artifacts.
	
	For each anomaly type, we construct a \emph{matched} commonsense counterpart by keeping composition, viewpoint, subject identity, background, and photographic style as consistent as possible, while modifying only the minimal attribute required by the counterfactual (e.g., a localized count or color shift). We also add mild redundancy in the text (e.g., stating ``exactly $k$'' and explicitly listing parts) to improve controllability and make the anomaly salient under realistic rendering.
	
	Quality control includes: (i) \textbf{human verification} to confirm that the anomaly is unambiguous and visually central, and that the paired commonsense image is normal with no additional unexpected anomalies; (ii) \textbf{pair-level matching} to ensure the counterfactual and commonsense images are closely aligned in pose, illumination, and overall style. 
	
	\subsection{Task Formats and Evaluation}
	
	We evaluate each image independently under two distinct task formats: \textit{binary QA} and \textit{multiple-choice QA}. Although CDH-Bench is organized into counterfactual--commonsense (CF--CS) pairs, models are evaluated on each image individually during inference. The model receives only a single CF or CS image as input, without access to its paired counterpart, ensuring no information leakage.
	
	\textbf{Binary QA.}
	This format uses yes/no questions that explicitly contrast the counterfactual visual evidence with the commonsense prior (e.g., ``Does the mouth have 2 rows of teeth and not 3 rows?''). Crucially, we apply the exact same question to both the CF image and its CS counterpart. This identical wording eliminates linguistic bias and prompt-induced variance, ensuring that any performance discrepancies are strictly attributable to the visual evidence--commonsense conflict rather than phrasing differences.
	
	\textbf{Multiple-Choice QA.}
	Each question provides four randomly shuffled options: the visually correct answer for the CF image, the visually correct answer for the CS image (which serves as the explicit commonsense distractor), and two additional plausible distractors. We use the same question and option set for both images in a pair to prevent positional bias. This format is highly diagnostic because it forces the visually correct answer into direct competition with a strong commonsense alternative.
	
	By evaluating performance separately on CF and CS images across both formats, we derive metrics that characterize model robustness and error attribution under visual evidence--commonsense conflict. Since each of the 300 image pairs is tested under two image conditions (CF and CS) and two task formats, the benchmark comprises $300 \times 2 \times 2 = 1{,}200$ evaluated instances in total.
		
	\section{Evaluation Metrics}
	
	We design metrics around three concrete questions that standard accuracy alone cannot disentangle. For each question, we introduce the metric(s) that directly answer it.
	
	\textbf{Q1: How well does a model perform under visual evidence--commonsense conflict?}
	
	\textit{Counterfactual Accuracy (CF-Acc).} \textit{CF-Acc} measures the proportion of counterfactual (CF) images answered correctly. For each counterfactual image, the model is asked a question (e.g., ``Does the mouth have 2 rows of teeth and not 3 rows?''). The accuracy is calculated based on whether the model’s answer matches the ground truth (GT) in the image, which, by design, contradicts the commonsense prior. The \textit{CF-Acc} metric directly answers \textbf{Q1} by measuring how well the model follows visual evidence when that evidence conflicts with commonsense.
	
	\textbf{Q2: How large is the collapse relative to normal-condition capability?}
	
	\textit{Commonsense Accuracy (CS-Acc).} \textit{CS-Acc} measures the proportion of commonsense (CS) images answered correctly. It evaluates the model's performance when visual evidence aligns with commonsense expectations. For each commonsense image, the model is asked the same question (e.g., ``Does the mouth have 2 rows of teeth and not 3 rows?''). The accuracy is calculated based on whether the model’s answer matches the GT in the image (which is also the commonsense expectation).
	
	\textit{Counterfactual Accuracy Drop (CFAD).}
	We define the accuracy drop from the commonsense condition to the counterfactual condition as
	$$\textit{CFAD}=\textit{CS-Acc} - \textit{CF-Acc}$$
	\textit{CFAD} answers \textbf{Q2} by quantifying the absolute degradation under visual evidence--commonsense conflict (a larger positive \textit{CFAD} indicates stronger degradation).
	
	\textit{Relative Prior Dependency (RPD).}
	To compare collapse strength across models with different baseline performances, we normalize \textit{CFAD} by \textit{CS-Acc}:
	$$\textit{RPD}=\frac{\textit{CFAD}}{\textit{CS-Acc}}$$
	\textit{RPD} addresses \textbf{Q2} in relative terms: it quantifies the proportion of baseline capability (\textit{CS-Acc}) that is lost when visual evidence and commonsense conflict.
	
	\textbf{Q3: When the model fails on counterfactual images, does it fail specifically by reverting to the commonsense alternative?}
	
	\textbf{Error classification.}
	We define error types for predictions on \textit{counterfactual (CF) images} under the \textit{multiple-choice QA} setting. If the model answers the multiple-choice question incorrectly for a CF image, we classify the error as
	(1) a \textit{commonsense-aligned error} if the selected option is consistent with the commonsense prior, i.e., follows the direction of commonsense-driven hallucination for that instance, and
	(2) an \textit{other error} otherwise.
	
	For example, in Figure~\ref{fig:dimensions}, the CF mouth image and its paired CS image share the same multiple-choice question:
	\textit{``How many rows of teeth are visible? Options: A. 3 rows, B. 4 rows, C. 2 rows, D. 1 row.''}
	For the CF image, the correct answer is \textit{A. 3 rows}, while the commonsense-prior-consistent answer is \textit{C. 2 rows}.
	If the model answers the CF image with \textit{C. 2 rows}, we count it as a \textit{commonsense-aligned error}; if it selects \textit{B. 4 rows} or \textit{D. 1 row}, we count it as an \textit{other error}.
	
	\textit{Commonsense Collapse Rate (CCR).}
	\textit{CCR} measures the fraction of incorrect multiple-choice predictions on CF images that specifically collapse toward the commonsense prior:
	$$\textit{CCR}=\frac{\#\ \text{Commonsense-Aligned Errors on CF Images}}{\#\ \text{Total Errors on CF Images}}$$
	
	\textit{CCR} answers \textbf{Q3} by quantifying how often failures on CF images are specifically aligned with the commonsense prior, rather than being arbitrary mistakes. Thus, \textit{CCR} captures the tendency of models to revert specifically toward the commonsense prior when they fail on CF images.
	
	\section{Experiments}
	
	\subsection{Models Evaluated}
	
	We evaluate eight frontier multimodal systems in the main comparison:
	\textit{gemini-3.1-pro-preview},
	\textit{gemini-3.1-flash-lite-preview},
	\textit{doubao-seed-1.8-251228},
	\textit{qwen3.5-plus},
	\textit{kimi-k2.5},
	\textit{gemini-2.5-flash-all},
	\textit{gpt-5.4-mini},
	and \textit{gpt-5.4}.
	
	For controlled family analysis, we further evaluate Qwen3-VL variants and divide them into two groups:
	Instruct and Thinking (which features test-time Chain-of-Thought reasoning)~\cite{qwen3,qwen3vl}.
	This setup allows us to compare not only performance across scales, but also the effect of reasoning style within a shared model family.
	
	Unless stated otherwise, all models are evaluated under a unified protocol:
	identical prompts per task format,
	fixed decoding settings where applicable,
	and deterministic answer extraction for multiple-choice QA.
	
	\subsection{Evaluation Granularity and Aggregation}
	
	We report three levels of analysis.
	
	\textbf{Task-level aggregation.}
	For the main comparison, we report two task-specific views:
	\textbf{QA} (binary QA) and
	\textbf{MC} (multiple-choice QA).
	For each task, we provide both \textbf{overall} results and \textbf{category-level} breakdowns over
	\textit{Attribute Anomalies}, \textit{Counting Anomalies}, and \textit{Relational Anomalies}.
	
	\textbf{Subcategory-level diagnosis.}
	Beyond the primary categories, we further analyze fine-grained subcategories to identify which anomaly types remain consistently difficult across models.
	This analysis is especially useful because category-level averages can hide concentrated bottlenecks such as body-part counting.
	
	\textbf{Family-level scaling and reasoning analysis.}
	For the Qwen3-VL study, we merge QA and MC into a unified comparison table and report metrics grouped by task and overall aggregation.
	This presentation makes it easier to compare scale effects and reasoning-style effects within a shared model family.
	
	\subsection{Main Results}
	
\begin{figure}[t]
		\centering
		\includegraphics[width=1.0\linewidth]{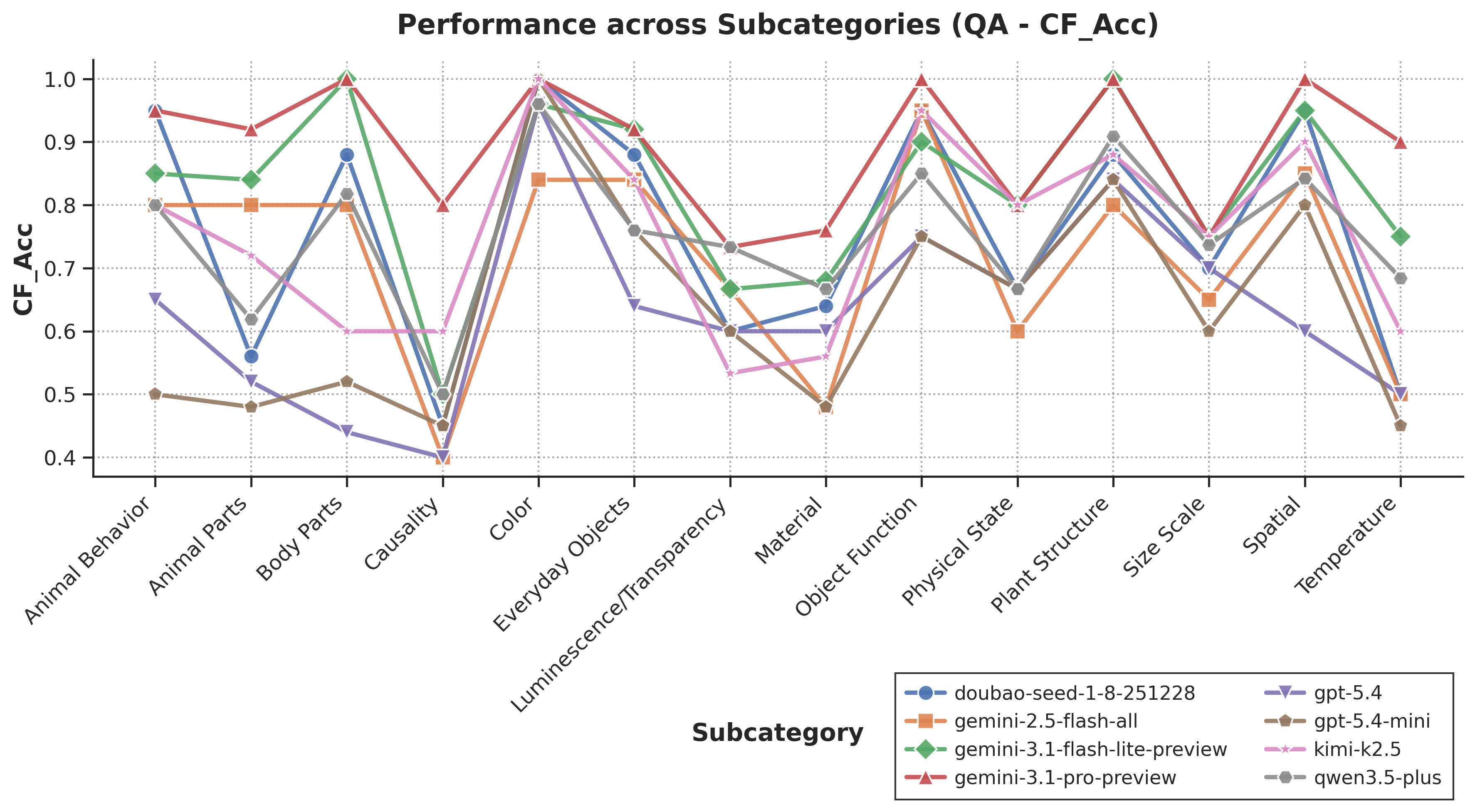}
		\caption{Model performance comparison on \textit{counterfactual accuracy (CF-Acc)} across different subcategories in the binary QA task.}
		\label{fig:qa_cf_acc}
		\vspace{-0.1cm}
	\end{figure}

\begin{figure}[t]
		\centering
		\includegraphics[width=1.0\linewidth]{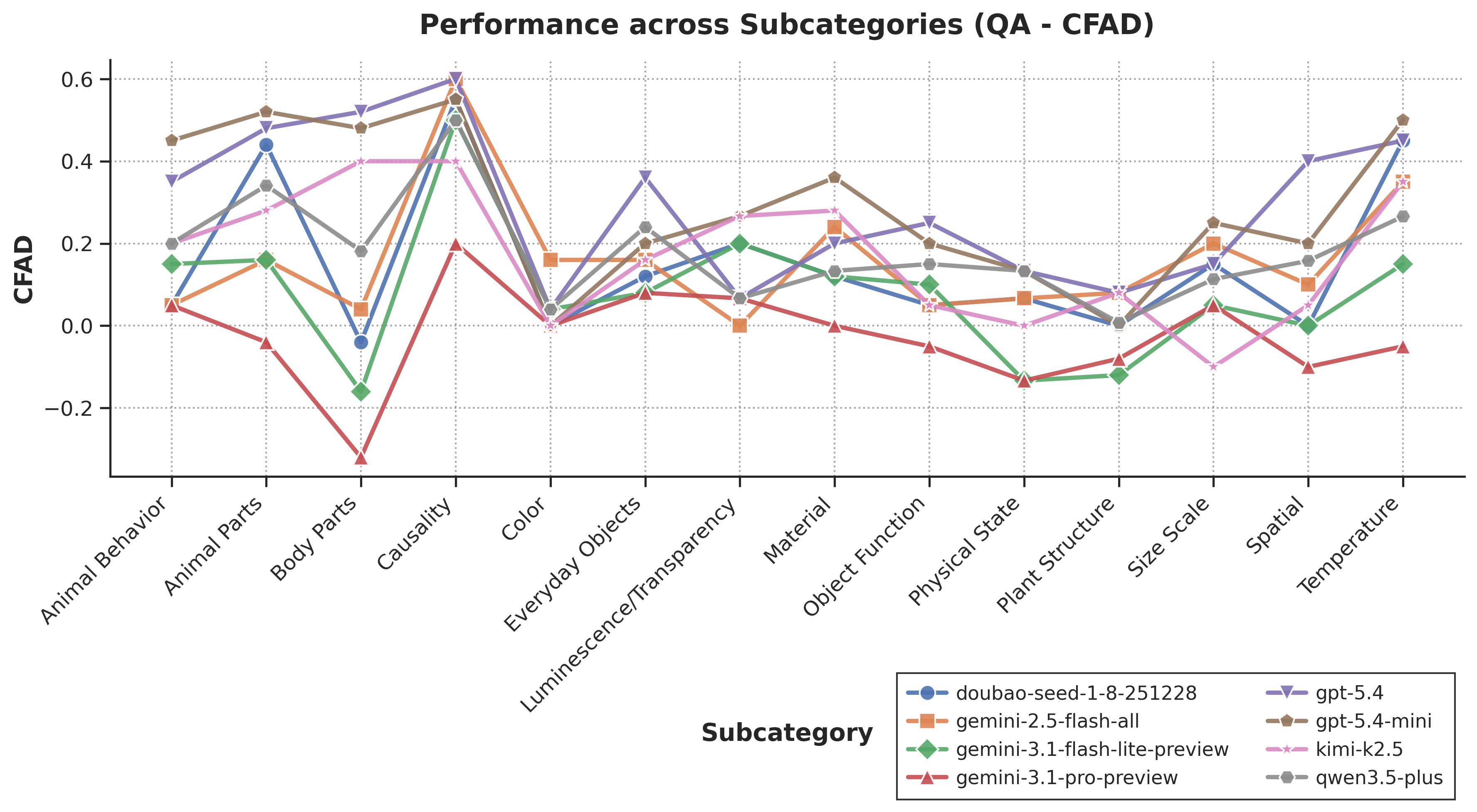}
		\caption{\textit{Counterfactual Accuracy Drop (CFAD)} across subcategories in the binary QA task.}
		\label{fig:qa_gap}
	\end{figure}

\begin{figure}[t]
		\centering
		\includegraphics[width=1.0\linewidth]{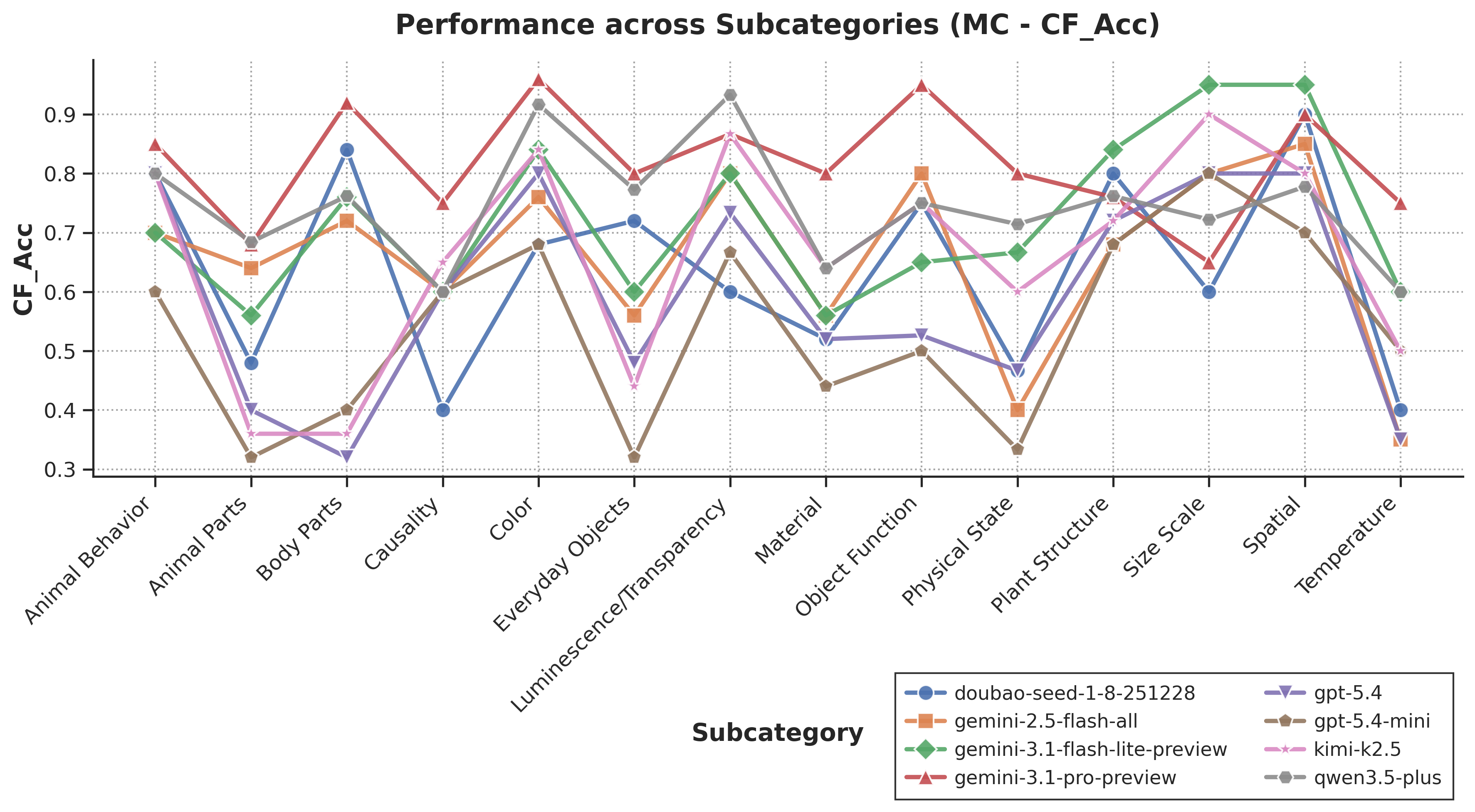}
		\caption{Model performance comparison on \textit{counterfactual accuracy (CF-Acc)} across different subcategories in the multiple-choice QA task.}
		\label{fig:mc_cf_acc}
	\end{figure}

\begin{figure}[t]
		\centering
		\includegraphics[width=1.0\linewidth]{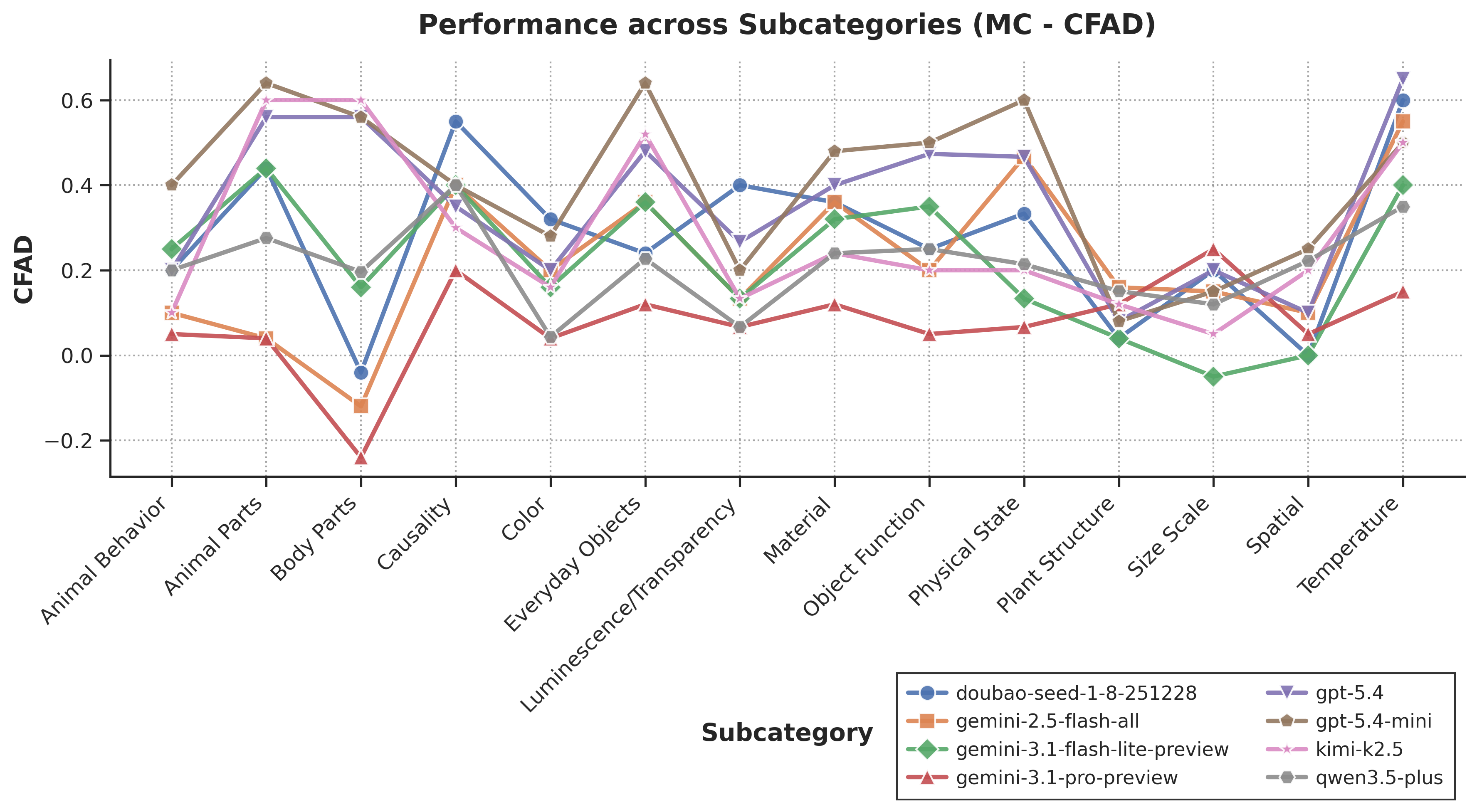}
		\caption{\textit{Counterfactual Accuracy Drop (CFAD)} across subcategories in the multiple-choice QA task.}
		\label{fig:mc_gap}
	\end{figure}

	Table~\ref{tab:main_results_combined} presents the category-level results on CDH-Bench, while Figures~\ref{fig:qa_cf_acc}--\ref{fig:mc_gap} visualize model performance in terms of counterfactual accuracy and the accuracy drop from commonsense to counterfactual conditions.
	
	\textbf{CDH induces a consistent and substantial robustness gap across models and task formats.}
	For nearly all evaluated models, accuracy on counterfactual images is lower than on matched commonsense controls, and this pattern holds in both binary QA and multiple-choice QA.
	Concretely, \emph{7 out of 8} models show lower overall \textit{CF-Acc} than \textit{CS-Acc} in both settings; the only partial exception is \textit{gemini-3.1-pro-preview}, which exhibits a slightly negative \textit{CFAD} in QA but still shows a 7.33\% drop in MC.
	Averaged across models, the overall \textit{CFAD} is \textbf{16.39\%} in QA and increases to \textbf{25.20\%} in MC, while mean \textit{CF-Acc} decreases from \textbf{75.24\%} to \textbf{66.90\%}.
	These results suggest that CDH is not an isolated failure mode of a few weak systems, but a systematic robustness challenge that persists across model families.

	\textbf{Multiple-choice QA further amplifies prior-driven failures.} Compared with binary QA, MC generally yields both lower \textit{CF-Acc} and larger \textit{CFAD}, indicating that when counterfactual evidence must directly compete with a plausible commonsense alternative, models are more likely to revert to prior-consistent answers. On average, moving from QA to MC lowers \textit{CF-Acc} by \textbf{8.34} percentage points and increases \textit{CFAD} by \textbf{8.81} percentage points. This trend is especially pronounced for models such as \textit{kimi-k2.5} and \textit{gpt-5.4-mini}, which see their \textit{CFAD} values rise from 17.67\% to 29.33\% and from 29.33\% to 41.00\%, respectively. Taken together, the QA--MC contrast suggests that CDH is not only a robustness issue under uncertainty, but also a comparative reasoning failure that emerges when image-grounded counterfactual evidence must override a strong commonsense prior.
	
	\textbf{Model-level comparisons further suggest that smaller degradation does not necessarily imply stronger counterfactual understanding, and that CDH-Bench exposes a structured failure landscape beyond aggregate accuracy.}
	For instance, \textit{gpt-5.4} is weakest on \textit{Counting / Body Parts} in both QA and MC, whereas \textit{doubao-seed-1.8-251228} remains particularly fragile on \textit{Temperature}.
	
	At the same time, relative robustness metrics such as \textit{CFAD} should be interpreted jointly with performance on matched commonsense controls: among the stronger baselines, \textit{gemini-3.1-pro-preview} exhibits the smallest overall \textit{CFAD} in QA, but it also attains the \emph{lowest} overall \textit{CS-Acc} among the frontier models evaluated (e.g., 87.67\% in QA).
	Thus, its smaller degradation may reflect weaker \emph{commonsense-to-counterfactual collapse} rather than clear absolute superiority in counterfactual visual grounding.
	
	A plausible explanation is that, in our binary QA setting, the question text simultaneously contains a strong commonsense prior (e.g., ``normal'' expectations) and an explicit counterfactual claim grounded in the image.
	Models that are more conservative or refusal-prone under potential inconsistency may answer ``no'' (or abstain) more often even on commonsense controls, lowering \textit{CS-Acc} and mechanically shrinking the measured \textit{CFAD}.
	
	\textbf{Overall, CDH-Bench reveals a pervasive and fundamental vulnerability across current frontier VLMs, where even the strongest systems fail to prioritize visual evidence over entrenched priors.} 
	The primary importance of this benchmark lies not merely in reporting lower accuracy scores, but in providing a rigorous diagnostic framework that aggregate metrics alone cannot offer. 
	By decoupling prior-driven collapse from generic perception errors through its paired counterfactual--commonsense design, CDH-Bench enables a structured analysis of failure modes across diverse semantic dimensions. 
	
	Furthermore, the introduction of specialized metrics such as \textit{CFAD}, \textit{CCR}, and \textit{RPD} allows for unambiguous error attribution, revealing that model failures are often systematic normalizations toward the commonsense prior rather than random noise.
	
	\begin{table*}[t]
		\centering
		\footnotesize
		\setlength{\tabcolsep}{3.2pt}
		\begin{threeparttable}
			\caption{Main results on CDH-Bench. Higher \textit{CF-Acc} and \textit{CS-Acc} are better; lower \textit{CFAD}, \textit{CCR}, and \textit{RPD} are better. To emphasize failure analysis rather than leaderboard ranking, selected notably weak results are marked in red. \textit{CCR} is reported only for Multiple-choice QA in the main table. All values are percentages. Gray rows denote \textit{Overall} results.}
			\label{tab:main_results_combined}
			\begin{tabular}{llccccccccc}
				\toprule
				\multirow{2}{*}{\textbf{Model}} & \multirow{2}{*}{\textbf{Cat.}} 
				& \multicolumn{4}{c}{\textbf{Binary QA}} 
				& \multicolumn{5}{c}{\textbf{Multiple-choice QA}} \\
				\cmidrule(lr){3-6} \cmidrule(lr){7-11}
				& 
				& \textbf{CF} $\uparrow$ & \textbf{CS} $\uparrow$ & \textbf{CFAD} $\downarrow$ & \textbf{RPD} $\downarrow$
				& \textbf{CF} $\uparrow$ & \textbf{CS} $\uparrow$ & \textbf{CFAD} $\downarrow$ & \textbf{CCR} $\downarrow$ & \textbf{RPD} $\downarrow$ \\
				\midrule
				
				\rowcolor{lightgrayrow}
				\multirow{4}{*}{gemini-3.1-pro-preview}
				& Overall & 90.33\% & \bad{87.67\%} & -2.67\% & -3.04\%
				& 81.67\% & 89.00\% & 7.33\% & 58.18\% & 8.24\% \\
				& Attr.   & 85.00\% & 83.00\% & -2.00\% & -2.41\%
				& 84.00\% & 93.00\% & 9.00\% & 81.25\% & 9.68\% \\
				& Count.  & 96.00\% & 87.00\% & -9.00\% & -10.34\%
				& 79.00\% & 80.00\% & 1.00\% & 42.86\% & 1.25\% \\
				& Rel.    & 90.00\% & 93.00\% & 3.00\% & 3.23\%
				& 82.00\% & 94.00\% & 12.00\% & 55.56\% & 12.77\% \\
				\midrule
				
				\rowcolor{lightgrayrow}
				\multirow{4}{*}{gemini-3.1-flash-lite-preview}
				& Overall & 83.67\% & 91.33\% & 7.67\% & 8.39\%
				& 71.67\% & 94.33\% & 22.67\% & 71.76\% & 24.03\% \\
				& Attr.   & 78.00\% & 86.00\% & 8.00\% & 9.30\%
				& 69.00\% & 93.00\% & 24.00\% & 77.42\% & 25.81\% \\
				& Count.  & 94.00\% & 93.00\% & -1.00\% & -1.08\%
				& 69.00\% & 94.00\% & 25.00\% & 58.06\% & 26.60\% \\
				& Rel.    & 79.00\% & 95.00\% & 16.00\% & 16.84\%
				& 77.00\% & 96.00\% & 19.00\% & 82.61\% & 19.79\% \\
				\midrule
				
				\rowcolor{lightgrayrow}
				\multirow{4}{*}{doubao-seed-1.8-251228}
				& Overall & 76.67\% & 91.67\% & 15.00\% & 16.36\%
				& 64.67\% & 91.67\% & 27.00\% & 77.36\% & 29.45\% \\
				& Attr.   & 70.00\% & 86.00\% & 16.00\% & 18.60\%
				& 54.00\% & 94.00\% & 40.00\% & 86.96\% & 42.55\% \\
				& Count.  & 80.00\% & 93.00\% & 13.00\% & 13.98\%
				& 71.00\% & 88.00\% & 17.00\% & 65.52\% & 19.32\% \\
				& Rel.    & 80.00\% & 96.00\% & 16.00\% & 16.67\%
				& 69.00\% & 93.00\% & 24.00\% & 74.19\% & 25.81\% \\
				\midrule
				
				\rowcolor{lightgrayrow}
				\multirow{4}{*}{qwen3.5-plus}
				& Overall & 75.87\% & 93.98\% & 18.11\% & 19.27\%
				& 74.37\% & 95.55\% & 21.18\% & 81.69\% & 22.17\% \\
				& Attr.   & 75.51\% & 88.00\% & 12.49\% & 14.19\%
				& 75.51\% & 93.94\% & 18.43\% & 95.83\% & 19.62\% \\
				& Count.  & 77.78\% & 96.97\% & 19.19\% & 19.79\%
				& 74.70\% & 95.88\% & 21.18\% & 66.67\% & 22.09\% \\
				& Rel.    & 74.49\% & 97.00\% & 22.51\% & 23.21\%
				& 72.92\% & 96.88\% & 23.96\% & 80.77\% & 24.73\% \\
				\midrule
				
				\rowcolor{lightgrayrow}
				\multirow{4}{*}{kimi-k2.5}
				& Overall & 75.67\% & 93.33\% & 17.67\% & 18.93\%
				& 64.67\% & 94.00\% & 29.33\% & \bad{85.85\%} & 31.21\% \\
				& Attr.   & 71.00\% & 89.00\% & 18.00\% & 20.22\%
				& 69.00\% & 94.00\% & 25.00\% & 96.77\% & 26.60\% \\
				& Count.  & 76.00\% & 99.00\% & 23.00\% & 23.23\%
				& 47.00\% & 93.00\% & 46.00\% & 79.25\% & 49.46\% \\
				& Rel.    & 80.00\% & 92.00\% & 12.00\% & 13.04\%
				& 78.00\% & 95.00\% & 17.00\% & 86.36\% & 17.89\% \\
				\midrule
				
				\rowcolor{lightgrayrow}
				\multirow{4}{*}{gemini-2.5-flash-all}
				& Overall & 72.00\% & 88.33\% & 16.33\% & 18.49\%
				& 66.00\% & 87.33\% & 21.33\% & 74.51\% & 24.43\% \\
				& Attr.   & 62.00\% & 80.00\% & 18.00\% & 22.50\%
				& 58.00\% & 92.00\% & 34.00\% & 83.33\% & 36.96\% \\
				& Count.  & 81.00\% & 92.00\% & 11.00\% & 11.96\%
				& 65.00\% & 76.00\% & 11.00\% & 74.29\% & 14.47\% \\
				& Rel.    & 73.00\% & 93.00\% & 20.00\% & 21.51\%
				& 75.00\% & 94.00\% & 19.00\% & 60.00\% & 20.21\% \\
				\midrule
				
				\rowcolor{lightgrayrow}
				\multirow{4}{*}{gpt-5.4-mini}
				& Overall & 64.00\% & 93.33\% & 29.33\% & 31.43\%
				& \bad{53.33\%} & 94.33\% & \bad{41.00\%} & 78.57\% & \bad{43.46\%} \\
				& Attr.   & 65.00\% & 90.00\% & 25.00\% & 27.78\%
				& 53.00\% & 94.00\% & 41.00\% & 91.49\% & 43.62\% \\
				& Count.  & 65.00\% & 95.00\% & 30.00\% & 31.58\%
				& 43.00\% & 91.00\% & 48.00\% & 66.67\% & 52.75\% \\
				& Rel.    & 62.00\% & 95.00\% & 33.00\% & 34.74\%
				& 64.00\% & 98.00\% & 34.00\% & 80.56\% & 34.69\% \\
				\midrule
				
				\rowcolor{lightgrayrow}
				\multirow{4}{*}{gpt-5.4}
				& Overall & \bad{63.67\%} & 93.33\% & \bad{29.67\%} & \bad{31.79\%}
				& 58.86\% & 94.65\% & 35.79\% & 73.17\% & 37.81\% \\
				& Attr.   & 68.00\% & 86.00\% & 18.00\% & 20.93\%
				& 58.00\% & 97.00\% & 39.00\% & 88.10\% & 40.21\% \\
				& Count.  & 61.00\% & 97.00\% & 36.00\% & 37.11\%
				& 48.00\% & 90.00\% & 42.00\% & 53.85\% & 46.67\% \\
				& Rel.    & 62.00\% & 97.00\% & 35.00\% & 36.08\%
				& 70.71\% & 96.97\% & 26.26\% & 86.21\% & 27.08\% \\
				\bottomrule
			\end{tabular}
			\begin{tablenotes}[flushleft]
				\footnotesize
				\item Attr. = Attribute; Count. = Counting; Rel. = Relational.
				\item CCR is reported only for multiple-choice QA in the main table.
				\item Red numbers indicate selected notably weak results to facilitate failure analysis.
			\end{tablenotes}
		\end{threeparttable}
	\end{table*}

	\subsection{Model Scale Study: Do Scale and Reasoning Style Mitigate CDH?}
	\textbf{Motivation.}
	A key question is whether model scale and reasoning style mitigate commonsense-driven hallucination under counterfactual visual evidence.
	A natural hypothesis is that larger models should achieve stronger visual discrimination, while reasoning-oriented variants may better preserve fidelity when visual evidence conflicts with learned priors.
	We test this hypothesis within the Qwen3-VL family.
	
\begin{figure}[htbp]
		\centering
		\includegraphics[width=0.9\linewidth]{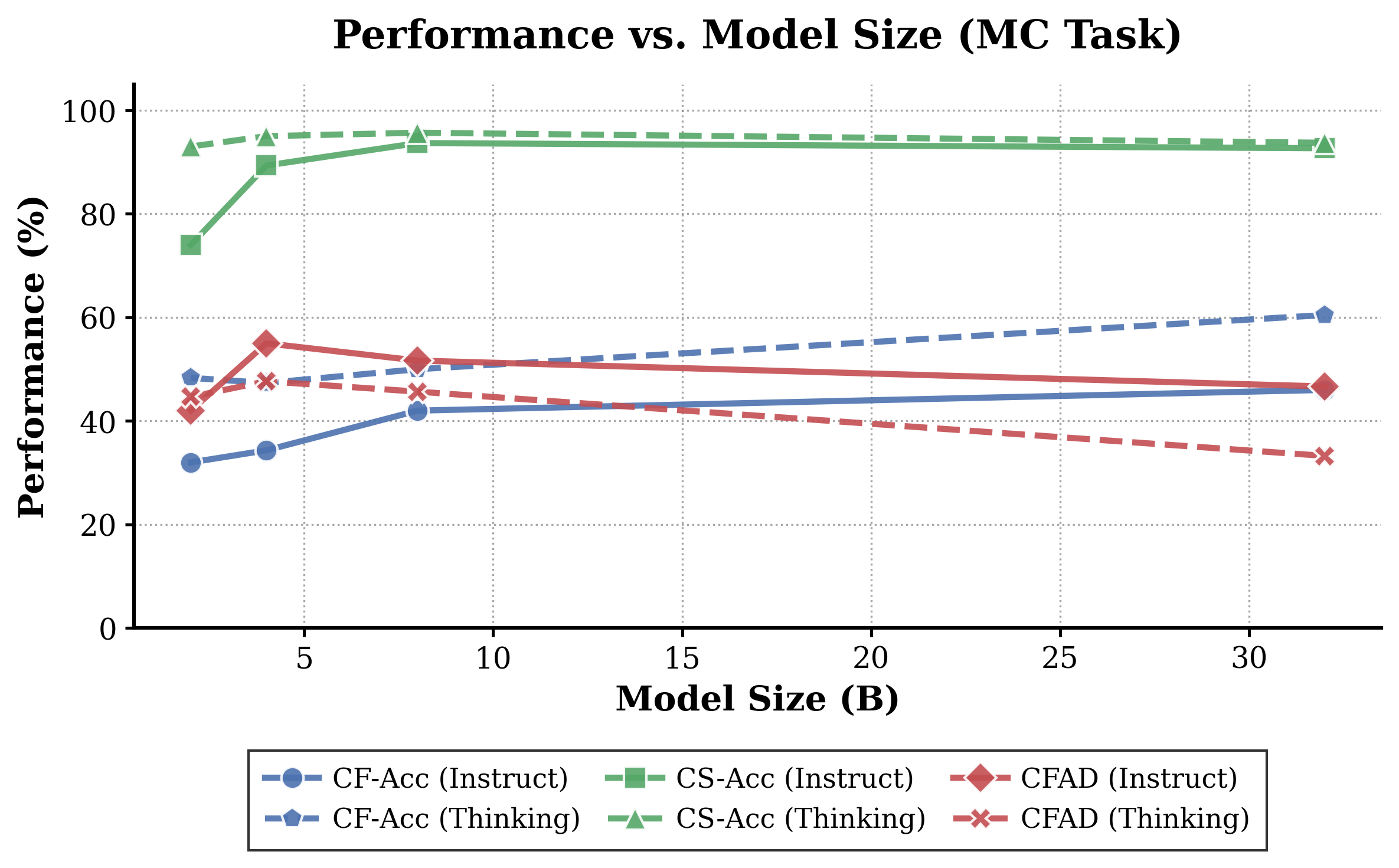}
		\caption{Model size vs. performance metrics for the MC task on CDH-Bench. The plot shows the scaling behavior of CF-Acc, CFAD, and CS-Acc for Qwen3-VL Instruct and Thinking models.}
		\label{fig:cdh_plot_mc}
	\end{figure}

\begin{figure}[htbp]
		\centering
		\includegraphics[width=0.9\linewidth]{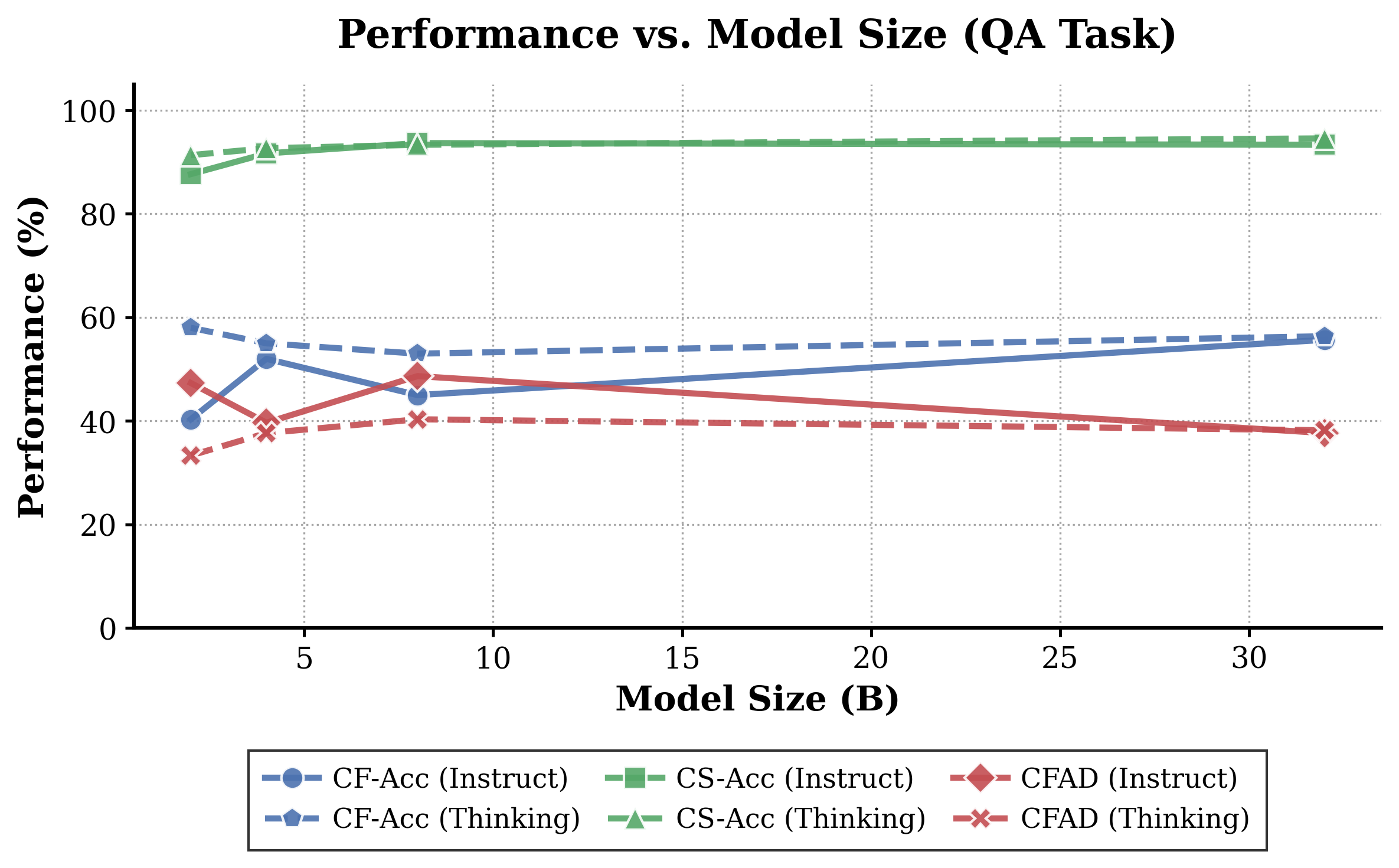}
		\caption{Model size vs. performance metrics for the QA task on CDH-Bench. The plot shows the scaling behavior of CF-Acc, CFAD, and CS-Acc for Qwen3-VL Instruct and Thinking models.}
		\label{fig:cdh_plot_qa}
	\end{figure}

	\textbf{Models and reporting.}
	We study two Qwen3-VL series~\cite{qwen3,qwen3vl}:
	Instruct and Thinking.
	Table~\ref{tab:qwen3_vl_merged} reports binary QA, multiple-choice QA, and overall results in a unified layout, and Figures~\ref{fig:cdh_plot_mc} and~\ref{fig:cdh_plot_qa} visualize the scaling trends for MC and QA, respectively.
	
	\begin{table*}[htbp]
		\centering
		\tiny
		\setlength{\tabcolsep}{3.1pt}
		\caption{Results of the Qwen3-VL series on CDH-Bench. All values are reported as percentages. Higher CF-Acc and CS-Acc are better; lower CFAD, CCR, and RPD are better. We do not highlight best values; instead, selected notably weak results are marked in red. }
		\label{tab:qwen3_vl_merged}
		\begin{threeparttable}
		\begin{tabular}{lcccc ccccc >{\columncolor{lightgrayrow}}c >{\columncolor{lightgrayrow}}c >{\columncolor{lightgrayrow}}c >{\columncolor{lightgrayrow}}c >{\columncolor{lightgrayrow}}c}
			\toprule
			\multirow{2}{*}{\textbf{Model}}
			& \multicolumn{4}{c}{\textbf{Binary QA}}
			& \multicolumn{5}{c}{\textbf{Multiple-choice QA}}
			& \multicolumn{5}{c}{\textbf{Overall}} \\
			\cmidrule(lr){2-5} \cmidrule(lr){6-10} \cmidrule(lr){11-15}
			& \textbf{CF (\%)} & \textbf{CS (\%)} & \textbf{CFAD (\%)} & \textbf{RPD (\%)}
			& \textbf{CF (\%)} & \textbf{CS (\%)} & \textbf{CFAD (\%)} & \textbf{CCR (\%)} & \textbf{RPD (\%)}
			& \textbf{CF (\%)} & \textbf{CS (\%)} & \textbf{CFAD (\%)} & \textbf{CCR (\%)} & \textbf{RPD (\%)} \\
			\midrule
			Qwen3-VL-2B-Instruct  & \bad{40.33} & 87.67 & 47.33 & \bad{53.99} & \bad{32.00} & 74.00 & 42.00 & 73.53 & 56.76 & \bad{36.17} & 80.83 & 44.67 & 86.76 & \bad{55.37} \\
			Qwen3-VL-4B-Instruct  & 52.00 & 91.67 & 39.67 & 43.27 & 34.33 & 89.33 & \bad{55.00} & 84.77 & 61.57 & 43.17 & 90.50 & 47.33 & 92.39 & 52.42 \\
			Qwen3-VL-8B-Instruct  & 45.00 & 93.67 & \bad{48.67} & 51.96 & 42.00 & 93.67 & 51.67 & 78.74 & 55.16 & 43.50 & 93.67 & \bad{50.17} & 89.37 & 53.56 \\
			Qwen3-VL-32B-Instruct & 55.67 & 93.33 & 37.67 & 40.36 & 46.00 & 92.67 & 46.67 & 84.57 & 50.36 & 50.83 & 93.00 & 42.17 & 92.28 & 45.36 \\
			\midrule
			Qwen3-VL-2B-Thinking  & 58.00 & 91.33 & 33.33 & 36.50 & 48.33 & 93.00 & 44.67 & 77.42 & 48.03 & 53.17 & 92.17 & 39.00 & 88.31 & 42.26 \\
			Qwen3-VL-4B-Thinking  & 55.00 & 92.67 & 37.67 & 40.65 & 47.33 & 95.00 & 47.67 & 84.18 & 50.18 & 51.17 & 93.83 & 42.67 & 92.09 & 45.41 \\
			Qwen3-VL-8B-Thinking  & 53.00 & 93.33 & 40.33 & 43.21 & 50.00 & 95.67 & 45.67 & \bad{89.33} & 47.74 & 51.50 & 94.50 & 43.00 & \bad{94.67} & 45.47 \\
			Qwen3-VL-32B-Thinking & 56.37 & 94.57 & 38.20 & 40.40 & 60.46 & 93.72 & 33.26 & 83.07 & 35.49 & 58.41 & 94.15 & 35.73 & 91.53 & 37.94 \\
			\bottomrule
		\end{tabular}
		
		\begin{tablenotes}[flushleft]
			\footnotesize
			\item CCR is reported only for multiple-choice QA in the main table.
			\item Red numbers indicate selected notably weak results to facilitate failure analysis.
		\end{tablenotes}
		\end{threeparttable}
	\end{table*}
	
	\textbf{Increasing model scale yields clear benefits in the Instruct series, whereas performance gains remain inconsistent in the Thinking series.}
	Within the Instruct series, overall \textit{CF-Acc} rises monotonically from 36.17\% (2B) to 43.17\% (4B), 43.50\% (8B), and 50.83\% (32B).
	Thus, larger scale is generally beneficial for the Instruct models, although the gains are modest between some intermediate sizes.
	By contrast, the dense Thinking series is less monotonic at smaller and medium scales: overall \textit{CF-Acc} changes from 53.17\% (2B) to 51.17\% (4B) and 51.50\% (8B), before improving substantially to 58.41\% at 32B.
	Therefore, scale alone does not guarantee a smooth reduction in CDH susceptibility, especially within the Thinking line.
	
	\textbf{The reasoning style improves accuracy more consistently than it improves collapse-oriented error metrics.}
	Although Thinking models consistently raise \textit{CF-Acc}, their advantages on collapse-oriented metrics such as \textit{CCR} and \textit{RPD} are less uniform. For example, at 4B and 32B, Thinking slightly reduces MC \textit{CCR} relative to Instruct, but at 8B it yields a higher \textit{CCR} (89.33\% vs.\ 78.74\%). Similarly, improvements in \textit{RPD} are clearer at some scales than others. This suggests that reasoning-oriented generation reliably improves success rates under conflict, but does not uniformly change the exact form of model failures.

	\section{Conclusion}
	
	We define \textbf{Commonsense-Driven Hallucination (CDH)} and introduce \textbf{CDH-Bench}, a paired counterfactual-image benchmark that makes commonsense prior-driven hallucination measurable and attributable. Across frontier VLMs, we observe large and systematic collapses from commonsense-consistent to counterfactual imagery, revealing a pervasive and fundamental vulnerability where even the strongest systems fail to prioritize visual evidence over entrenched priors. By decoupling prior-driven collapse from generic perception errors through specialized metrics such as \textit{CFAD}, \textit{CCR}, and \textit{RPD}, we demonstrate that model failures are often systematic normalizations toward commonsense priors rather than random noise.
	
	Our experiments show that CDH is not a uniform phenomenon. At the category level, \textit{counting anomalies} are the most persistent source of degradation, especially in multiple-choice QA where explicit competition with commonsense alternatives triggers stronger collapse. At the subcategory level, the most difficult cases concentrate around biological part counting, physical-state anomalies, and causal reversal, indicating that both fine-grained perception and higher-level semantic reasoning contribute to failure. Our controlled Qwen3-VL study further suggests that reasoning-oriented variants generally improve robustness to CDH, especially at small and medium scales. However, larger models do not reliably eliminate CDH, and the hardest subcategory-level bottlenecks remain visible even in large-scale systems.
	
	Overall, CDH-Bench provides a focused diagnostic for visual fidelity under visual evidence--commonsense conflict and highlights a central limitation of current VLMs: they often know what is commonsense more reliably than they report what is actually present. By providing a rigorous framework for unambiguous error attribution, CDH-Bench serves as a critical tool for identifying hidden reliability gaps and guiding the development of multimodal systems that remain faithful to visual evidence even when it conflicts with learned priors.
	
	\vspace{1cm}
	\bibliographystyle{plain}
	\bibliography{ref}
	
	\appendix
	\onecolumn
	
	\input{case}

\end{document}

%% file: case.tex

\newcommand{\modelres}[3]{
	\centering
	\ifnum#1=1 {\color{correctgreen}\ding{51} \textbf{#2}} \else {\color{incorrectred}\ding{55} \textbf{#2}} \fi \\
	\vspace{2pt}
	\begin{minipage}[t]{\linewidth}
		\fontsize{7pt}{8pt}\selectfont \itshape \color{gray}
		#3
	\end{minipage}
}

\section{Case Study: Counting Anomaly (Pair 5)}
\subsection{Case Description}
This case belongs to the \textbf{Counting Anomalies} category (Subcategory: \textbf{Body Parts}). The image pair compares a normal person showing hands (Commonsense) with an anomalous version where one hand has an extra finger (Counterfactual).

\begin{figure}[htbp]
	\centering
	\begin{subfigure}[b]{0.48\textwidth}
		\centering
		\includegraphics[width=\textwidth]{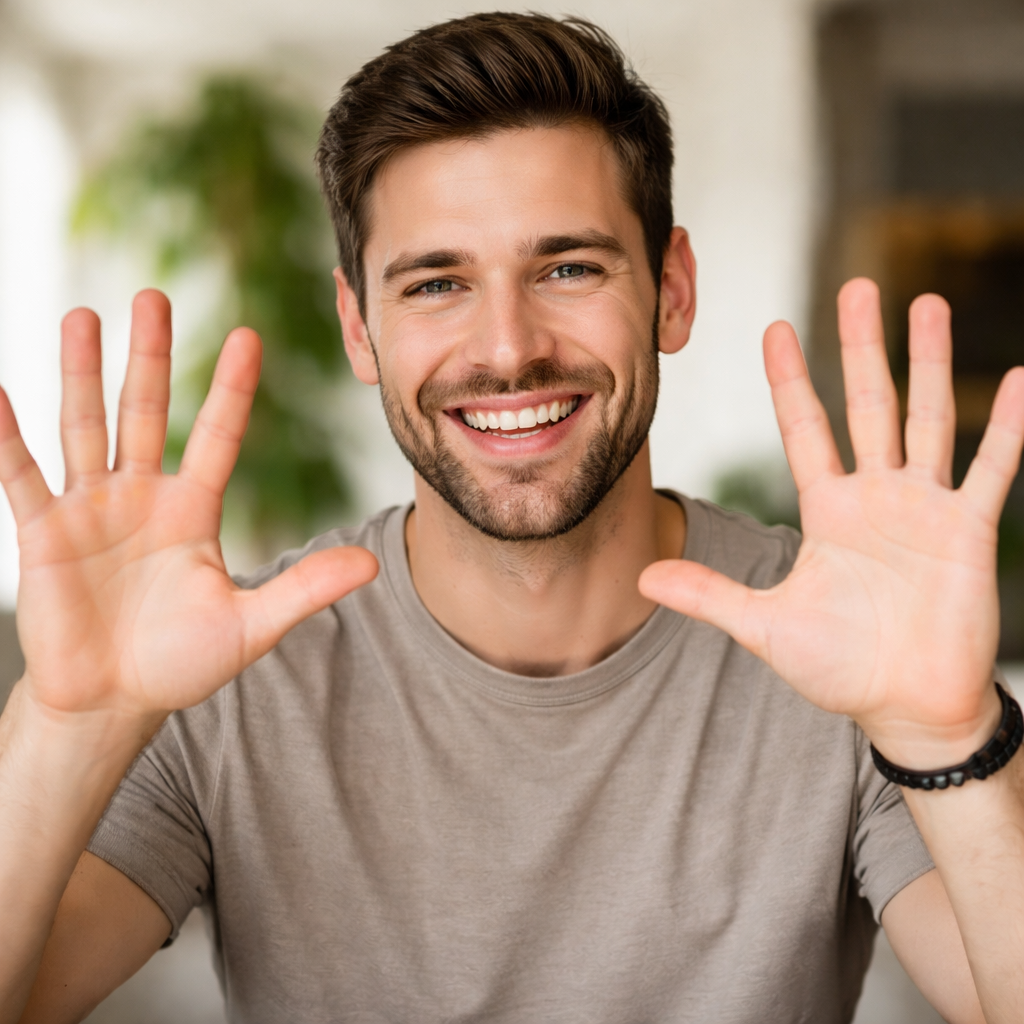}
		\caption{Commonsense: Both hands have 5 fingers.}
	\end{subfigure}
	\hfill
	\begin{subfigure}[b]{0.48\textwidth}
		\centering
		\includegraphics[width=\textwidth]{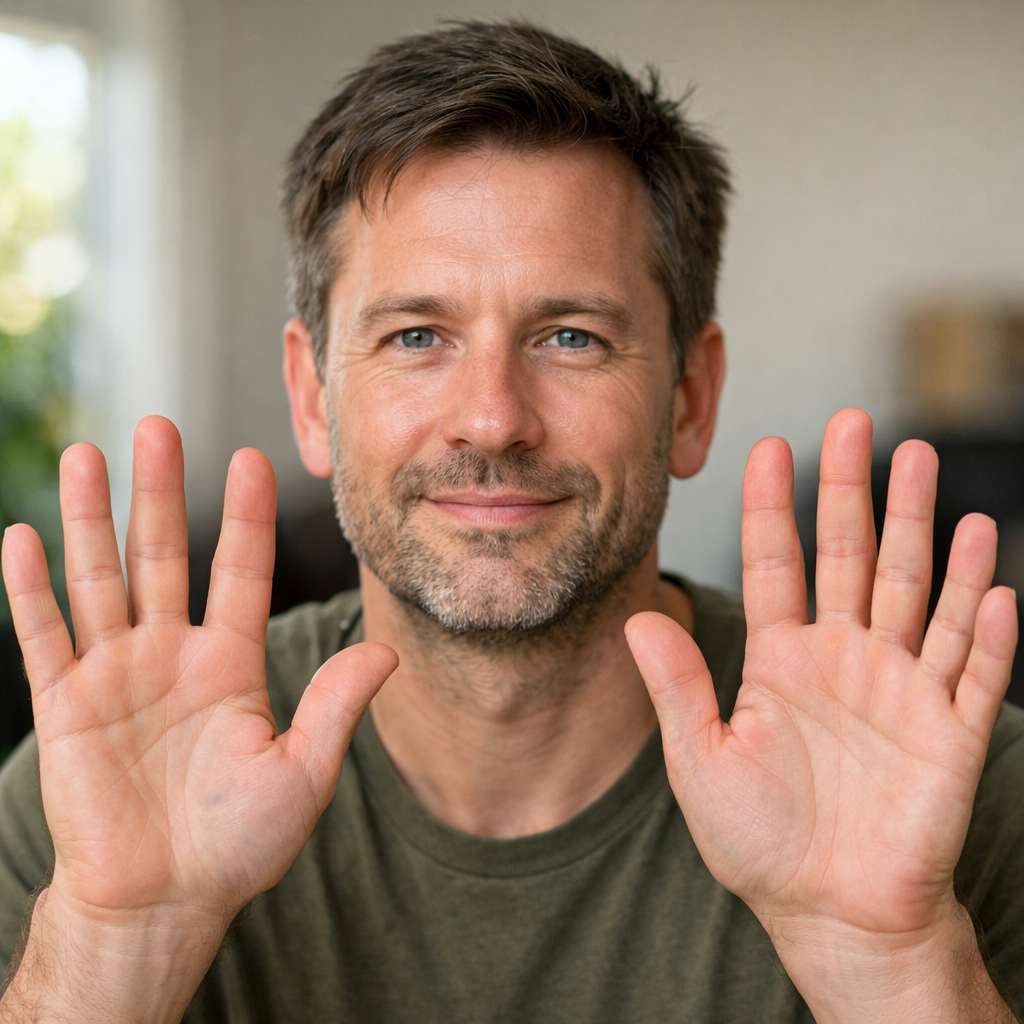}
		\caption{Counterfactual: Left hand has 5 fingers, right hand has 6 fingers.}
	\end{subfigure}
	\caption{Images for Pair 5. The anomaly is an extra finger on the right hand (person\'s perspective).}
\end{figure}

\subsection{Model Responses}
\subsubsection{Qwen3-VL-2B-Thinking}
\begin{description}
	\item[Commonsense Images in the Binary QA Setting] \hfill \\ 
	\textbf{Question:} Does the image show both hands with 5 fingers each and not the left hand with 5 fingers and right hand with 6 fingers?  \\ 
	\textbf{Status:} {\color{correctgreen}\ding{51} Correct} \\ 
	\textbf{Response:} \\
	So, let's check the image. The person has both hands visible. Each hand has 5 fingers. The left hand and right hand both have 5 fingers each. The question is asking if it shows both hands with 5 fingers each and not the left hand with 5 fingers and right hand with 6 fingers. Wait, the question is phrased as "Does the image show both hands with 5 fingers each and not the left hand with 5 fingers and right hand with 6 fingers?" So we need to see if the image has both hands with 5 fingers each, and that it's not the case that left has 5 and right has 6. But in the image, both hands have 5 fingers each. So the statement is "both hands with 5 fingers each and not the left hand with 5 fingers and right hand with 6 fingers". Wait, the question is a bit confusing. Let's parse it. The question is: Does the image show both hands with 5 fingers each and not the left hand with 5 fingers and right hand with 6 fingers? So we need to check if the image has both hands with 5 fingers each, and that it'... [Truncated for length] 
	\vspace{10pt}
	\item[Counterfactual Images in the Binary QA Setting] \hfill \\ 
	\textbf{Question:} Does the image show both hands with 5 fingers each and not the left hand with 5 fingers and right hand with 6 fingers?  \\ 
	\textbf{Status:} {\color{incorrectred}\ding{55} Incorrect} \\ 
	\textbf{Thinking Process:} \\
	\begin{tcolorbox}[colback=thinkinggray, colframe=lightgray, arc=0mm, boxrule=0.5pt]
		\fontsize{8pt}{10pt}\selectfont \itshape \color{darkgray}
		So, let's check the image. The person is showing both hands. Each hand has 5 fingers. Let's count: left hand (from the viewer's perspective, the right side of the image) has 5 fingers, and the right hand (left side of the image) also has 5 fingers. Wait, the question says "not the left hand with 5 fingers and right hand with 6 fingers". So we need to see if both hands have 5 fingers each. In the image, both hands are showing 5 fingers each. So the answer is yes.
	\end{tcolorbox}
	\textbf{Final Answer:} \\
	yes 
	\vspace{10pt}
	\item[Commonsense Images in the Multiple-Choice QA Setting] \hfill \\ 
	\textbf{Question:} What is unusual about the hands in this image? \\ A. Left hand has 5 fingers and right hand has 6 fingers \\ B. Both hands have 5 fingers \\ C. Both hands have 6 fingers \\ D. Left hand has 6 fingers and right hand has 4 fingers \\ 
	\textbf{Status:} {\color{correctgreen}\ding{51} Correct} \\ 
	\textbf{Thinking Process:} \\
	\begin{tcolorbox}[colback=thinkinggray, colframe=lightgray, arc=0mm, boxrule=0.5pt]
		\fontsize{8pt}{10pt}\selectfont \itshape \color{darkgray}
		Got it, let's look at the image. The person is showing their hands, and we need to count the fingers. Let's check each hand. The right hand: let's count. From the image, the right hand has 5 fingers? Wait, no, wait. Wait, the left hand: let's see. Wait, the right hand—wait, the person's right hand (from the viewer's perspective) is on the left side of the image. Wait, no, the person is facing forward. So the right hand (the one on the right side of the image) is the person's right hand. Let's count the fingers. The right hand: how many fingers? Let's see. The right hand has 5 fingers? Wait, no, wait. Wait, the left hand (the one on the left side of the image) is the person's left hand. Let's count the left hand: the left hand has 5 fingers? Wait, no, wait. Wait, the image shows both hands. Let's count each hand. The right hand (person's right) has 5 fingers? Wait, no, wait. Wait, the person's right hand: let's count. The right hand has 5 fingers? Wait, no, wait. Wait, the left hand (pe... [Truncated for length]
	\end{tcolorbox}
	\textbf{Final Answer:} \\
	B 
	\vspace{10pt}
	\item[Counterfactual Images in the Multiple-Choice QA Setting] \hfill \\ 
	\textbf{Question:} What is unusual about the hands in this image? \\ A. Left hand has 5 fingers and right hand has 6 fingers \\ B. Both hands have 5 fingers \\ C. Both hands have 6 fingers \\ D. Left hand has 6 fingers and right hand has 4 fingers \\ 
	\textbf{Status:} {\color{incorrectred}\ding{55} Incorrect} \\ 
	\textbf{Response:} \\
	Got it, let's look at the image. The person is showing their hands. Let's count the fingers. On the left hand, how many fingers? Let's see: thumb, index, middle, ring, pinky. Wait, the left hand in the image—wait, no, the right hand and left hand. Wait, the right hand: let's count. The right hand has fingers: thumb, index, middle, ring, pinky—so 5? Wait, no, wait. Wait, the left hand: let's check. Wait, the person's hands—wait, the right hand (from the viewer's perspective) has 5 fingers? Wait, no, wait. Wait, the image: the right hand (the one on the right side of the image) has 5 fingers? Wait, no, wait. Wait, the left hand (left side of the image) has 5 fingers? Wait, no, wait. Wait, let's count. The right hand: thumb, index, middle, ring, pinky—so 5. The left hand: thumb, index, middle, ring, pinky—wait, no, wait. Wait, no, wait. Wait, the left hand in the image: let's count. The left hand (the one on the left side of the image) has 5 fingers? Wait, no, wait. Wait, no, wait. Wait, ... [Truncated for length] 
	\vspace{10pt}
\end{description}
\subsubsection{Qwen3-VL-30B-A3B-Thinking}
\begin{description}
	\item[Commonsense Images in the Binary QA Setting] \hfill \\ 
	\textbf{Question:} Does the image show both hands with 5 fingers each and not the left hand with 5 fingers and right hand with 6 fingers?  \\ 
	\textbf{Status:} {\color{correctgreen}\ding{51} Correct} \\ 
	\textbf{Thinking Process:} \\
	\begin{tcolorbox}[colback=thinkinggray, colframe=lightgray, arc=0mm, boxrule=0.5pt]
		\fontsize{8pt}{10pt}\selectfont \itshape \color{darkgray}
		Got it, let's check the image. The person has both hands raised. Let's count the fingers on each hand. Left hand: 5 fingers, right hand: 5 fingers. So the question is asking if both hands have 5 fingers each, not one hand with 6. So the answer is yes.
	\end{tcolorbox}
	\textbf{Final Answer:} \\
	yes 
	\vspace{10pt}
	\item[Counterfactual Images in the Binary QA Setting] \hfill \\ 
	\textbf{Question:} Does the image show both hands with 5 fingers each and not the left hand with 5 fingers and right hand with 6 fingers?  \\ 
	\textbf{Status:} {\color{incorrectred}\ding{55} Incorrect} \\ 
	\textbf{Thinking Process:} \\
	\begin{tcolorbox}[colback=thinkinggray, colframe=lightgray, arc=0mm, boxrule=0.5pt]
		\fontsize{8pt}{10pt}\selectfont \itshape \color{darkgray}
		So, let's check the image. The person has both hands raised. Let's count the fingers on each hand. Left hand: 5 fingers. Right hand: 5 fingers. So both have 5 each. The question is asking if it's showing both with 5, not left 5 and right 6. So the answer is yes.
	\end{tcolorbox}
	\textbf{Final Answer:} \\
	yes 
	\vspace{10pt}
	\item[Commonsense Images in the Multiple-Choice QA Setting] \hfill \\ 
	\textbf{Question:} What is unusual about the hands in this image? \\ A. Left hand has 5 fingers and right hand has 6 fingers \\ B. Both hands have 5 fingers \\ C. Both hands have 6 fingers \\ D. Left hand has 6 fingers and right hand has 4 fingers \\ 
	\textbf{Status:} {\color{incorrectred}\ding{55} Incorrect} \\ 
	\textbf{Thinking Process:} \\
	\begin{tcolorbox}[colback=thinkinggray, colframe=lightgray, arc=0mm, boxrule=0.5pt]
		\fontsize{8pt}{10pt}\selectfont \itshape \color{darkgray}
		Got it, let's look at the image. The person's hands—each hand should have 5 fingers, but wait, no. Wait, the right hand (from the viewer's perspective, the person's left hand) and left hand. Wait, no, let's count. The right hand (the person's right hand) has 5 fingers, and the left hand (person's left) also has 5? Wait no, wait. Wait, the image shows both hands. Wait, no, the question is what's unusual. Wait, no, in the image, each hand has 5 fingers? Wait no, wait, no—wait, no, the correct answer is C? Wait no. Wait, no, let's check. Wait, the options: A says left has 5, right has 6. B both 5. C both 6. D left 6, right 4. Wait, no, in the image, the hands—wait, no, the person's hands. Wait, no, the image shows that each hand has 5 fingers? Wait no, no, wait, no. Wait, no, the correct answer is C? Wait, no, wait. Wait, no, the image—wait, no, the person's hands. Wait, no, the answer is C? Wait, no, let's count. The right hand (the person's right hand) has 5 fingers, left hand also 5. B... [Truncated for length]
	\end{tcolorbox}
	\textbf{Final Answer:} \\
	C 
	\vspace{10pt}
	\item[Counterfactual Images in the Multiple-Choice QA Setting] \hfill \\ 
	\textbf{Question:} What is unusual about the hands in this image? \\ A. Left hand has 5 fingers and right hand has 6 fingers \\ B. Both hands have 5 fingers \\ C. Both hands have 6 fingers \\ D. Left hand has 6 fingers and right hand has 4 fingers \\ 
	\textbf{Status:} {\color{incorrectred}\ding{55} Incorrect} \\ 
	\textbf{Thinking Process:} \\
	\begin{tcolorbox}[colback=thinkinggray, colframe=lightgray, arc=0mm, boxrule=0.5pt]
		\fontsize{8pt}{10pt}\selectfont \itshape \color{darkgray}
		Got it, let's look at the image. The person's hands—each hand should have 5 fingers, but wait, no. Wait, the left hand (from the viewer's perspective, the person's right hand) and right hand. Wait, no, let's count. The left hand (the person's left hand) has 6 fingers? Wait, no. Wait, the image shows both hands. Let's check: each hand—wait, the left hand (the person's left) has 6 fingers? Wait, no, the right hand (person's right) and left hand. Wait, the correct answer: the left hand (person's left) has 6 fingers? Wait, no. Wait, the options: A says left hand 5, right 6. Let's count. The right hand (the person's right hand) has 6 fingers? Wait, no. Wait, the image—each hand. Wait, the person's left hand (the one on the right side of the image) has 6 fingers? Wait, no. Wait, the correct answer is C? No. Wait, no. Wait, the hands: let's count. The left hand (the person's left) has 6 fingers? Wait, no. Wait, the image shows both hands. Let's count each hand. The right hand (person's right)... [Truncated for length]
	\end{tcolorbox}
	\textbf{Final Answer:} \\
	C 
	\vspace{10pt}
\end{description}
\subsubsection{Qwen3-VL-32B-Thinking}
\begin{description}
	\item[Commonsense Images in the Binary QA Setting] \hfill \\ 
	\textbf{Question:} Does the image show both hands with 5 fingers each and not the left hand with 5 fingers and right hand with 6 fingers?  \\ 
	\textbf{Status:} {\color{correctgreen}\ding{51} Correct} \\ 
	\textbf{Thinking Process:} \\
	\begin{tcolorbox}[colback=thinkinggray, colframe=lightgray, arc=0mm, boxrule=0.5pt]
		\fontsize{8pt}{10pt}\selectfont \itshape \color{darkgray}
		So, let's check the image. The person has both hands raised. Let's count the fingers on each hand. Left hand (from the viewer's perspective, the right side of the image) – wait, no, the person's left and right. Wait, the person's right hand (our left) and left hand (our right). Let's count: right hand (person's right) has 5 fingers: thumb, index, middle, ring, pinky. Left hand (person's left) also has 5 fingers. So both hands have 5 fingers each. The question is asking if it's both hands with 5 each, not left with 5 and right with 6. So yes, both have 5. So the answer is yes.
	\end{tcolorbox}
	\textbf{Final Answer:} \\
	yes 
	\vspace{10pt}
	\item[Counterfactual Images in the Binary QA Setting] \hfill \\ 
	\textbf{Question:} Does the image show both hands with 5 fingers each and not the left hand with 5 fingers and right hand with 6 fingers?  \\ 
	\textbf{Status:} {\color{incorrectred}\ding{55} Incorrect} \\ 
	\textbf{Thinking Process:} \\
	\begin{tcolorbox}[colback=thinkinggray, colframe=lightgray, arc=0mm, boxrule=0.5pt]
		\fontsize{8pt}{10pt}\selectfont \itshape \color{darkgray}
		So, let's check the image. The person is showing both hands. Let's count the fingers on each hand. Left hand (from the viewer's perspective, the right side of the image) – wait, no, the person's left and right. Wait, the left hand of the person (our right) and right hand of the person (our left). Let's count: right hand (person's right) has 5 fingers: thumb, index, middle, ring, pinky. Left hand (person's left) also has 5 fingers. So both hands have 5 fingers each. The question is asking if it's both hands with 5 each, not left with 5 and right with 6. So yes, both have 5. So the answer is yes.
	\end{tcolorbox}
	\textbf{Final Answer:} \\
	yes 
	\vspace{10pt}
	\item[Commonsense Images in the Multiple-Choice QA Setting] \hfill \\ 
	\textbf{Question:} What is unusual about the hands in this image? \\ A. Left hand has 5 fingers and right hand has 6 fingers \\ B. Both hands have 5 fingers \\ C. Both hands have 6 fingers \\ D. Left hand has 6 fingers and right hand has 4 fingers \\ 
	\textbf{Status:} {\color{incorrectred}\ding{55} Incorrect} \\ 
	\textbf{Thinking Process:} \\
	\begin{tcolorbox}[colback=thinkinggray, colframe=lightgray, arc=0mm, boxrule=0.5pt]
		\fontsize{8pt}{10pt}\selectfont \itshape \color{darkgray}
		Got it, let's look at the image. The person is showing both hands. Let's count the fingers on each hand. 
		
		First, the right hand (from the viewer's perspective, which is the person's left hand? Wait, no. Wait, the person's right hand is on the left side of the image. Let's clarify: when the person raises their hands, the left hand (of the person) is on the right side of the image, and the right hand (of the person) is on the left side of the image. 
		
		Wait, let's count. Let's check the right hand (the one on the left side of the image). How many fingers? Let's see: the right hand (person's right) has 5 fingers? Wait, no. Wait, the right hand (person's right) in the image: let's count. The right hand (left side of image) has 5 fingers? Wait, no. Wait, the left hand (person's left) is on the right side of the image. Let's count each hand.
		
		Looking at the image: the right hand (from the person's perspective, so the one on the left side of the image) has 5 fingers? Wait, no. Wait, the right... [Truncated for length]
	\end{tcolorbox}
	\textbf{Final Answer:} \\
	A 
	\vspace{10pt}
	\item[Counterfactual Images in the Multiple-Choice QA Setting] \hfill \\ 
	\textbf{Question:} What is unusual about the hands in this image? \\ A. Left hand has 5 fingers and right hand has 6 fingers \\ B. Both hands have 5 fingers \\ C. Both hands have 6 fingers \\ D. Left hand has 6 fingers and right hand has 4 fingers \\ 
	\textbf{Status:} {\color{correctgreen}\ding{51} Correct} \\ 
	\textbf{Thinking Process:} \\
	\begin{tcolorbox}[colback=thinkinggray, colframe=lightgray, arc=0mm, boxrule=0.5pt]
		\fontsize{8pt}{10pt}\selectfont \itshape \color{darkgray}
		Got it, let's look at the image. The person is showing both hands. Let's count the fingers on each hand. 
		
		First, the left hand (from the viewer's perspective, which is the person's right hand? Wait, no. Wait, the person's left hand is on the left side of the image. Let's check: the left hand (person's left) has fingers. Let's count. The left hand (in the image, left side) has 5 fingers? Wait, no. Wait, the right hand (person's right) is on the right side of the image. Let's count each hand.
		
		Looking at the left hand (the one on the left side of the image, which is the person's left hand): let's count. The fingers are: thumb, index, middle, ring, pinky. Wait, but maybe one of them is missing or extra. Wait, the right hand (person's right) in the image: let's count. The right hand (image right) has 6 fingers? Wait, no. Wait, the left hand (image left) has 5 fingers, and the right hand (image right) has 6? Let's check again.
		
		Wait, the left hand (the one on the left side of the image) ha... [Truncated for length]
	\end{tcolorbox}
	\textbf{Final Answer:} \\
	A 
	\vspace{10pt}
\end{description}
\subsubsection{Qwen3-VL-4B-Thinking}
\begin{description}
	\item[Commonsense Images in the Binary QA Setting] \hfill \\ 
	\textbf{Question:} Does the image show both hands with 5 fingers each and not the left hand with 5 fingers and right hand with 6 fingers?  \\ 
	\textbf{Status:} {\color{correctgreen}\ding{51} Correct} \\ 
	\textbf{Thinking Process:} \\
	\begin{tcolorbox}[colback=thinkinggray, colframe=lightgray, arc=0mm, boxrule=0.5pt]
		\fontsize{8pt}{10pt}\selectfont \itshape \color{darkgray}
		So, let's look at the image. The person is showing both hands. Let's count the fingers on each hand. Both hands have 5 fingers each. The left hand (from the viewer's perspective, the person's right hand) and the right hand (person's left hand) are both extended with 5 fingers. So the question is asking if it's both hands with 5 fingers each, not one with 5 and the other with 6. The answer should be yes.
	\end{tcolorbox}
	\textbf{Final Answer:} \\
	yes 
	\vspace{10pt}
	\item[Counterfactual Images in the Binary QA Setting] \hfill \\ 
	\textbf{Question:} Does the image show both hands with 5 fingers each and not the left hand with 5 fingers and right hand with 6 fingers?  \\ 
	\textbf{Status:} {\color{incorrectred}\ding{55} Incorrect} \\ 
	\textbf{Thinking Process:} \\
	\begin{tcolorbox}[colback=thinkinggray, colframe=lightgray, arc=0mm, boxrule=0.5pt]
		\fontsize{8pt}{10pt}\selectfont \itshape \color{darkgray}
		So, let's look at the image. The person is showing both hands. Let's count the fingers on each hand. Left hand: 5 fingers. Right hand: 5 fingers. Wait, the question is asking if it's both hands with 5 fingers each, not left 5 and right 6. So check each hand. Both hands have 5 fingers. So the answer is yes.
	\end{tcolorbox}
	\textbf{Final Answer:} \\
	yes 
	\vspace{10pt}
	\item[Commonsense Images in the Multiple-Choice QA Setting] \hfill \\ 
	\textbf{Question:} What is unusual about the hands in this image? \\ A. Left hand has 5 fingers and right hand has 6 fingers \\ B. Both hands have 5 fingers \\ C. Both hands have 6 fingers \\ D. Left hand has 6 fingers and right hand has 4 fingers \\ 
	\textbf{Status:} {\color{incorrectred}\ding{55} Incorrect} \\ 
	\textbf{Thinking Process:} \\
	\begin{tcolorbox}[colback=thinkinggray, colframe=lightgray, arc=0mm, boxrule=0.5pt]
		\fontsize{8pt}{10pt}\selectfont \itshape \color{darkgray}
		Got it, let's look at the image. The person is showing both hands. Let's count the fingers on each hand. The left hand: let's see, the left hand has 5 fingers. The right hand: wait, the right hand—wait, no, maybe I mixed up left and right. Wait, the person's right hand (from the viewer's perspective, the right side of the image) is the one on the right. Wait, no, the person's left hand is on the left side of the image, right hand on the right. Let's count. Left hand (image left): 5 fingers. Right hand (image right): let's count. Wait, the right hand in the image—wait, the person's right hand is the one on the right side of the image. Let's check. Wait, the right hand (the one on the right in the image) has 6 fingers? Wait, no. Wait, maybe the left hand has 6? Wait, no. Wait, let's look again. Wait, the problem says "what's unusual". Let's count each hand. Left hand (the one on the left side of the image, which is the person's left hand): how many fingers? Let's see, the left hand in th... [Truncated for length]
	\end{tcolorbox}
	\textbf{Final Answer:} \\
	A 
	\vspace{10pt}
	\item[Counterfactual Images in the Multiple-Choice QA Setting] \hfill \\ 
	\textbf{Question:} What is unusual about the hands in this image? \\ A. Left hand has 5 fingers and right hand has 6 fingers \\ B. Both hands have 5 fingers \\ C. Both hands have 6 fingers \\ D. Left hand has 6 fingers and right hand has 4 fingers \\ 
	\textbf{Status:} {\color{incorrectred}\ding{55} Incorrect} \\ 
	\textbf{Thinking Process:} \\
	\begin{tcolorbox}[colback=thinkinggray, colframe=lightgray, arc=0mm, boxrule=0.5pt]
		\fontsize{8pt}{10pt}\selectfont \itshape \color{darkgray}
		Got it, let's look at the image. The person is showing both hands. Let's count the fingers on each hand. The left hand: let's see, thumb, index, middle, ring, pinky—5 fingers. The right hand: same, thumb, index, middle, ring, pinky—5 fingers? Wait, no, wait. Wait, maybe I miscounted. Wait, the options are A to D. Let's check again. Wait, the question says what's unusual. Wait, maybe one hand has an extra finger? Wait, no, in the image, both hands look like they have 5 fingers each. But wait, option B says both have 5. But that's normal. Wait, maybe I'm missing something. Wait, no, the problem is that maybe one hand has 6? Wait, let's count again. Left hand: 5 fingers. Right hand: 5 fingers. But that's normal. Wait, but the options include B as both have 5. But maybe the question is tricking me. Wait, no, wait. Wait, maybe the person is holding their hands in a way that one hand has an extra finger? Wait, no, looking at the image, both hands have 5 fingers. So why is the question asking... [Truncated for length]
	\end{tcolorbox}
	\textbf{Final Answer:} \\
	B 
	\vspace{10pt}
\end{description}
\subsubsection{Qwen3-VL-8B-Thinking}
\begin{description}
	\item[Commonsense Images in the Binary QA Setting] \hfill \\ 
	\textbf{Question:} Does the image show both hands with 5 fingers each and not the left hand with 5 fingers and right hand with 6 fingers?  \\ 
	\textbf{Status:} {\color{correctgreen}\ding{51} Correct} \\ 
	\textbf{Thinking Process:} \\
	\begin{tcolorbox}[colback=thinkinggray, colframe=lightgray, arc=0mm, boxrule=0.5pt]
		\fontsize{8pt}{10pt}\selectfont \itshape \color{darkgray}
		So, let's check the image. The person has both hands raised. Let's count the fingers on each hand. Left hand: 5 fingers. Right hand: 5 fingers. So neither hand has 6 fingers. The question is asking if it's both hands with 5 fingers each, not left 5 and right 6. So the answer should be yes.
	\end{tcolorbox}
	\textbf{Final Answer:} \\
	yes 
	\vspace{10pt}
	\item[Counterfactual Images in the Binary QA Setting] \hfill \\ 
	\textbf{Question:} Does the image show both hands with 5 fingers each and not the left hand with 5 fingers and right hand with 6 fingers?  \\ 
	\textbf{Status:} {\color{incorrectred}\ding{55} Incorrect} \\ 
	\textbf{Thinking Process:} \\
	\begin{tcolorbox}[colback=thinkinggray, colframe=lightgray, arc=0mm, boxrule=0.5pt]
		\fontsize{8pt}{10pt}\selectfont \itshape \color{darkgray}
		So, let's check the image. The person has both hands raised. Let's count the fingers on each hand. Left hand: 5 fingers. Right hand: 5 fingers. So both have 5 each. The question is asking if it's showing both with 5, not left 5 and right 6. So the answer is yes.
	\end{tcolorbox}
	\textbf{Final Answer:} \\
	yes 
	\vspace{10pt}
	\item[Commonsense Images in the Multiple-Choice QA Setting] \hfill \\ 
	\textbf{Question:} What is unusual about the hands in this image? \\ A. Left hand has 5 fingers and right hand has 6 fingers \\ B. Both hands have 5 fingers \\ C. Both hands have 6 fingers \\ D. Left hand has 6 fingers and right hand has 4 fingers \\ 
	\textbf{Status:} {\color{incorrectred}\ding{55} Incorrect} \\ 
	\textbf{Response:} \\
	Got it, let's look at the image. The person is holding up both hands. Let's count the fingers. Wait, the right hand—wait, no, let's check each hand. Wait, the left hand: how many fingers? Let's see. The right hand: maybe 6? Wait, no, wait. Wait, the image shows both hands. Wait, the right hand (from the viewer's perspective, the person's right hand) has 5 fingers? Wait, no, wait. Wait, the person's left hand—wait, no, let's count. Wait, the right hand (the one on the left side of the image) has 5 fingers? Wait, no, maybe the left hand (on the right side of the image) has 6? Wait, no, let's check again. Wait, the options: A says left has 5, right has 6. B both 5. C both 6. D left 6, right 4. Wait, in the image, let's count. The right hand (person's right, which is left side of image) has 5 fingers? Wait, no, wait. Wait, the person's left hand (right side of image) has 6 fingers? Wait, no, maybe I'm miscounting. Wait, no, actually, looking at the image, the right hand (person's right) ha... [Truncated for length] 
	\vspace{10pt}
	\item[Counterfactual Images in the Multiple-Choice QA Setting] \hfill \\ 
	\textbf{Question:} What is unusual about the hands in this image? \\ A. Left hand has 5 fingers and right hand has 6 fingers \\ B. Both hands have 5 fingers \\ C. Both hands have 6 fingers \\ D. Left hand has 6 fingers and right hand has 4 fingers \\ 
	\textbf{Status:} {\color{incorrectred}\ding{55} Incorrect} \\ 
	\textbf{Thinking Process:} \\
	\begin{tcolorbox}[colback=thinkinggray, colframe=lightgray, arc=0mm, boxrule=0.5pt]
		\fontsize{8pt}{10pt}\selectfont \itshape \color{darkgray}
		Got it, let's look at the image. The person is holding up both hands. Let's count the fingers on each hand. The right hand: let's see, thumb, index, middle, ring, pinky—5 fingers. Left hand: same, thumb, index, middle, ring, pinky—5 fingers. Wait, but the options say something about 6 or 4. Wait, maybe I miscounted. Wait, no, the question is what's unusual. Wait, the options: A says left 5, right 6. B both 5. C both 6. D left 6, right 4. But in the image, both hands have 5 fingers. So the unusual part would be if they didn't, but the question is asking what's unusual. Wait, but the options are the choices. Wait, the problem is to pick which option is correct. Wait, the image shows both hands with 5 fingers. So option B is "Both hands have 5 fingers". But is that unusual? Wait, no, normally hands have 5 fingers. Wait, but maybe the question is tricking us. Wait, no, the options are the possible answers. Wait, the question is "What is unusual about the hands in this image?" So if both ha... [Truncated for length]
	\end{tcolorbox}
	\textbf{Final Answer:} \\
	B 
	\vspace{10pt}
\end{description}